\documentclass[10pt,journal]{IEEEtran}

\usepackage{graphicx}
\usepackage{subfigure}
\usepackage[linesnumbered,ruled,vlined]{algorithm2e}
\usepackage{amsmath}
\usepackage{amssymb}
\usepackage{lipsum}
\usepackage{makecell} 
\usepackage{multirow}
\usepackage{booktabs,caption}
\usepackage{float}
\usepackage{lipsum}
\usepackage{dblfloatfix}
\usepackage{blindtext}
\usepackage{ragged2e}
\usepackage{xcolor}
\usepackage{array}
\usepackage{hyperref}

\usepackage{tabulary,amsfonts, adjustbox}

\usepackage[utf8x]{inputenc}
\usepackage{mathtools}
\newcolumntype{C}[1]{>{\centering\let\newline\\\arraybackslash\hspace{0pt}}m{#1}}

\ifCLASSINFOpdf
\else
\fi

\SetCommentSty{mycommfont}
\makeatletter
\DeclareRobustCommand{\oplustimes}{%
	\mathbin{\mathpalette\o@plus@times\relax}%
}
\newcommand{\o@plus@times}[2]{%
	\ooalign{$\m@th#1\oplus$\cr$\m@th#1\otimes$\cr}%
}
\SetKwInput{KwInput}{Input}                % Set the Input
\SetKwInput{KwOutput}{Output}              % set the Output
\begin{document}
	\title{Neuro-Symbolic Learning: Principles and Applications in Ophthalmology} 
	% a structure at a ~ so this ke.pdf an author's name from being broken across
	
	\author{Muhammad~Hassan$^1$, Haifei~Guan$^1$, Aikaterini~Melliou$^1$, Yuqi~Wang$^1$, Qianhui~Sun$^1$, Sen~Zeng$^1$, Wen~Liang$^1$, Yiwei~Zhang$^1$, Ziheng~Zhang$^1$, Qiuyue~Hu$^1$, Yang~Liu$^1$, Shunkai~Shi$^2$, Lin~An$^1$, Shuyue~Ma$^1$, Ijaz~Gul$^1$, Muhammad~Akmal~Rahee$^1$, Zhou~You$^7$, Canyang~Zhang$^1$, Vijay~Kumar~Pandey$^1$, Yuxing~Han$^1$, Yongbing~Zhang$^3$, Ming~Xu$^4$, Qiming~Huang$^4$, 	Jiefu~Tan$^5$, Qi~Xing$^5$,  
		Peiwu~Qin$^{1,*}$,  Dongmei~Yu$^{6,*}$% <-this % stops a space
		\thanks{*Corresponding author}
		\thanks{Muhammad~Hassan~(mhassandev@gmail.com)} 
		\thanks{Qin~Piewu (pwqin@sz.tsinghua.edu.cn)}
		\thanks{$^1$Shenzhen International Graduate School, Tsinghua University, China}
		\thanks{$^2$SDU-ANU Joint Science College, Shandong University, China}
		\thanks{$^3$Harbin Institute of Technology, Shenzhen, China}
		\thanks{$^4$Liwei Zhilian Company Ltd., China}
		\thanks{$^5$Weichao Zhineng Company Ltd., China}
		\thanks{$^6$School of Mechanical, Electrical \& Information Engineering, Shandong University, China}
		\thanks{$^7$College of Computer Science and Technology, Jilin University, China}
		% <-this % stops a space
	}

	\maketitle
	% As a general rule, do no\textsf{\textsf{}}t put math, special symbols or citations
	\begin{abstract}
		Neural networks have been rapidly expanding in recent years, with novel strategies and applications. However, challenges such as interpretability, explainability, robustness, safety, trust, and sensibility remain unsolved in neural network technologies, despite the fact that they will unavoidably be addressed for critical applications. Attempts have been made to overcome the challenges in neural network computing by representing and embedding domain knowledge in terms of symbolic representations. Thus, the neuro-symbolic learning (NeSyL) notion emerged, which incorporates aspects of symbolic representation and bringing common sense into neural networks (NeSyL). In domains where interpretability, reasoning, and explainability are crucial, such as video and image captioning, question-answering and reasoning, health informatics, and genomics, NeSyL has shown promising outcomes. This review presents a comprehensive survey on the state-of-the-art NeSyL approaches, their principles, advances in machine and deep learning algorithms, applications such as opthalmology, and most importantly, future perspectives of this emerging field. 
	\end{abstract}
	

	% \fi
	\IEEEpeerreviewmaketitle

	\section{Introduction}
	\IEEEPARstart{N}{euro-symbolic} learning (NeSyL) is a relatively new field that combines aspects of both symbolic learning and artificial intelligence (AI). NeSyL hybridizes good old-fashioned AI into the advanced machine learning (ML) and deep networks towards an exciting interpretable AI. NeSyLs have extended AI capabilities and outperformed state-of-the-art deep learning models with higher accuracy in different domains, specifically in medical imaging and video reasoning. Researchers in this field are interested in understanding how the brain learns and represents knowledge, and how this knowledge can be used to improve AI based systems. In this review, we first provide an overview of the current state of the field. Furthermore, we discuss some of the key challenges that need to be addressed in order to advance the field. Finally, we suggest  possible future directions to broaden the employment of NeSyL to distinct critical fields.
	\begin{figure*}[h!]
	\centering
	\includegraphics[width=1\textwidth]{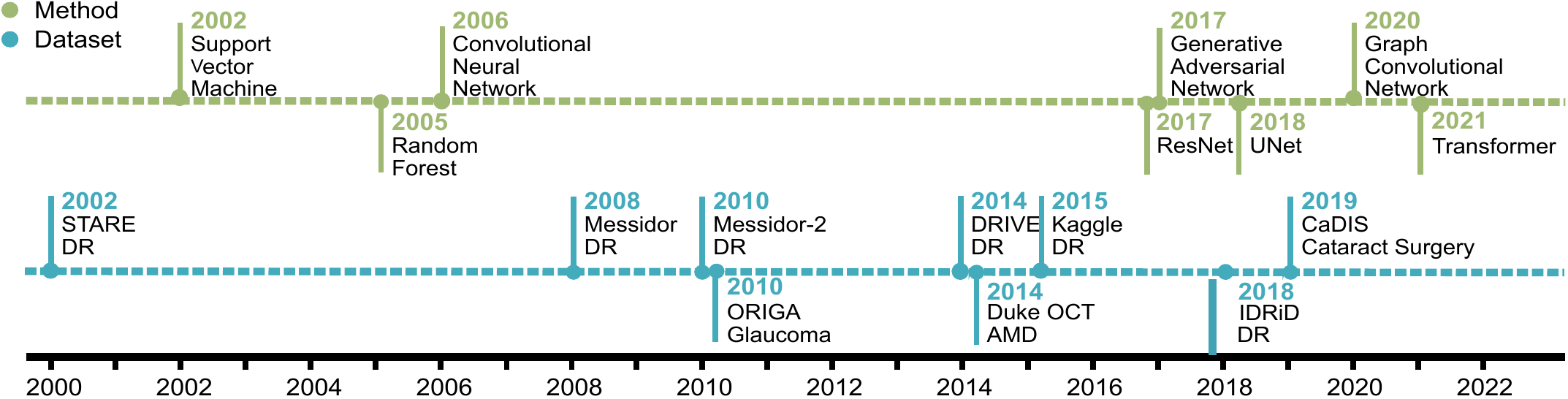}
	\caption{Development Timeline of machine learning and public datasets in the field of ophthalmology.  Machine learning methods such as SVM (2002) \cite{park2002active}, RF (2005) \cite{curnow2005multiplex}, and CNNs (2006)~\cite{checco2006cnn} have been widely used for ophthalmic disease classification, diagnosis, and prediction. With the rapid development of deep learning, network structures such as generative adversarial networks (2017) \cite{zisimopoulos2017can}, ResNet (2017) \cite{rattani2017fine}, UNet (2018) \cite{xiao2018weighted}, graph convolutional networks (2020) \cite{shi2020multi}, and transformer (2021) \cite{sun2021lesion} have been introduced to achieve more accurate results. Besides, the public availability of several ophthalmic datasets has significantly advanced the field of ophthalmic AI.}
	\label{fig:Time-vs-Dataset}
	\end{figure*}

\begin{figure}[t!]
		\centering
		\includegraphics[width=1\linewidth]{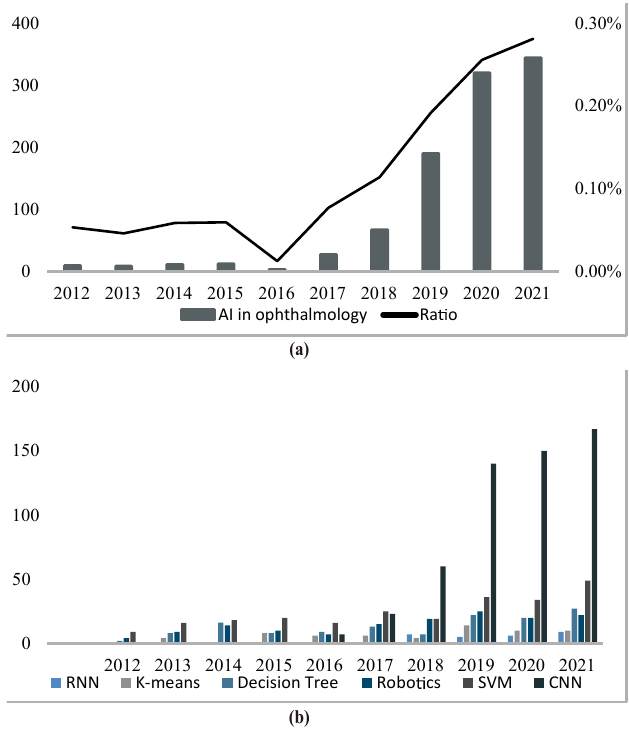}
		\caption{A comparative depiction of AI applications in opthalmology and different machine learning approaches in the last decade. The x-axis shows the percent-wise increase and the y-axis shows the yearly-wise division from 2012 to 2022. (a) AI applications in ophthalmology. (b) Applications of different state-of-the-art machine learning approaches.}
		\label{fig:AI-in-Opthamlogy}
		\end{figure}

	In clinical practices, AI has possessed increasing significance in the fields of ophthalmic diseases~\cite{gulshan2016development,ting2017development,mursch2020artificial} (Figure~\ref{fig:Time-vs-Dataset}), including screening, diagnosis, lesion segmentation, treatment, and prognosis. AI has a broad spectrum of applications in clinics and ophthalmology exemplifies in academic research and entrepreneurship. Thus, we use ophthalmology as example to illustrate the application of NeSyL in clinics. The application of AI in ophthalmology increased exponentially in the last decade (Figure~\ref{fig:AI-in-Opthamlogy}). AI in ophthalmology encompasses a wide range of ocular diseases such as cataracts, glaucoma, myopia, and diabetes retinopathy. The workflow of AI guided analysis of these diseases is composed of image classification, segmentation, and object detection and prediction. Although various AI algorithms have been adopted in the field of ophthalmology; however, deep learning (DL) is the dominant one. 

\begin{figure*}[h!]
		\centering
		\includegraphics[width=0.9\linewidth]{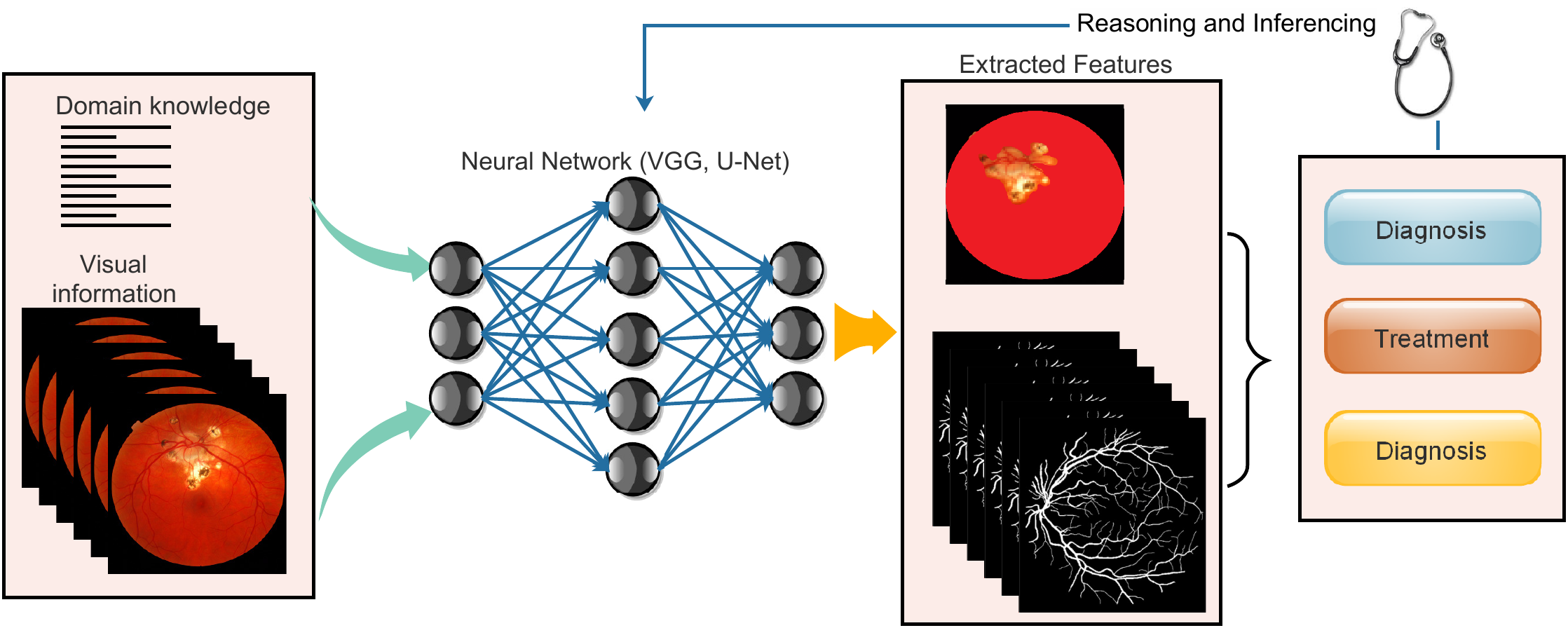}
		\caption{Workflow of DL approaches in ophthalmology.}
		\label{fig:DL-approaches-occular-disease}
	\end{figure*}

	We divide AI applications in ophthalmology into three tasks considering current progress. Then we discuss NeSyL approach to address the challenges in AI applications such as classification, segmentation, and detection into futuristic applications. The AI applications in opthalmology include, 1) the image classification and detection of ophthalmic disease for the examination of pathological severity; 2) image segmentation and partition of the retinal image into different parts according to their features and properties; and 3) disease prediction such as eyes related complications (Figure~\ref{fig:DL-approaches-occular-disease}).

	\subsection{Classification}
	Several DL approaches have been proposed to classify healthy and pathological cases based on image features in an end-to-end learning manner. The study \cite{bajwa2019two} used a two-stage approach to classify the retrieved disks as healthy or glaucoma patients. In the first step, the optic disc was localized and extracted from retinal fundus images using a convolutional neural network. In the second step, a deep convolutional network was applied to achieve the area under the curve (AUC) of 0.874, in which a 2.7\% improvement over the previous best results was observed \cite{fu2018joint}. Zhang et al. \cite{zhang2020attention} proposed a residual attention based multi-model ensemble method as an object identification network to extract the eyeball from the b-ultrasound image of the eye via ensembling three neural networks for cataract classification. The AUC was 0.975 and outperformed the counterpart approaches. 
	
	However, certain studies embedded additional information in terms of disease severity to categorize images. Wan et al. \cite{wan2018deep}, uses a Convolutional Neural Networks (CNNs)  model with transfer learning to perform diabetic retinopathy (DR) classification on the Kaggle, with a classification accuracy of 95.68\% and an AUC of 0.97. Peng et al. \cite{peng2019deepseenet} proposed the first patient-based scoring system, simulating the age-related macular degeneration (AMD) classification process by merging the De.pdfeeNet model with the age-related eye disease study (AREDS) simplified severity scale (0-5 points) of bilateral color fundus images.

	\subsection{Segmentation}
	The variation in retinal image patterns demands sophisticated DL segmentation models. For instance, Yan et al.~\cite{yan2018joint} proposed a U-Net like autoencoder for vessel segmentation which, combines segment and pixel level losses. Such a framework with two branches learns distinguishable features for vessel segmentation compared to monolithic loss based model. Akram et al.~\cite{akram2019adversarial} proposed generative adversarial network for retinal blood vessel segmentation with satisfactory results in terms of accuracy, F1-score and AUC. However, the method may perform poorly on thin vessel segmentation. For optic disc (OD) segmentation, Fu et al.~\cite{fu2018joint} developed M-Net to solve the OD and optic cup (OC) segmentation jointly in a one-stage multi-labels framework. The M-Net achieved significant results given multiscale input and side-output layers as compared to single-scale networks. Al-Bander et. [11] used multiscale sequential CNNs for simultaneous detection of fovea and OD on both Messidor~\cite{decenciere2014feedback} and Kaggle datasets~\cite{lardinois2017google}. The proposed method is less sensitive to preprocessing and does not need of vessel segmentation or border localization in order to detect the OD and foveal centers. A region proposal and cascaded network were proposed by Huang et al.~\cite{huang2020efficient} to realize robust and effective OD detection and fovea localization.
	
	\subsection{Prediction}
	The ML application has also been extended to the estimation of ocular disease. Among others, refractive error is a leading cause of visual impairment \cite{han2021myopia}. Traditional ML algorithms, such as Random Forest (RF) and Support Vector Machine (SVM) were used to predict the onset or development of myopia through numerical data like spherical equivalence (SE), and age~\cite{yang2020prediction,lin2018prediction}, while deep neural network (DNN) was employed to predict the progress of refractive error using retinal fundus images \cite{varadarajan2018deep}. Gordon et al. \cite{gordon2020evaluation} applied Conventional Cox proportional hazards regression modeling to assess the risk of glaucoma progression with untreated ocular hypertension and evaluated the influence of long-term intra-ocular pressure (IOP). Jiang et al. \cite{jiang2018predicting} utilized a deep temporal sequence, which extracts features using CNN and explores temporal information via long short term memory (LSTM) to predict the progression of posterior capsular opacification. Apart from the progression, predicting the outcome of ophthalmic therapy is another regular application of ML. Wolf-Dieter Vogl et al. \cite{vogl2017analyzing} applied a longitudinal mixed effects model to predict the outcome of anti-VEGF therapy for patients with exudative macular diseases.
	
	\subsection{AI Limitations and NeSyL }
	Despite the wide applications of AI in ocular image processing, there are limitations in terms of available datasets and ML models which hinder its implementation in clinical trials. Generally, annotated data obtained from hospitals are limited, specifically in segmentation or detection tasks, where the ground truths require annotations by experts. Sometimes the data is imbalanced with different quality and distributions. Some of the datasets are homogeneous in ethnic groups and lacking generalizability in real world applications. Medical data sets reflect true distributions in real clinical settings, but a minute imbalance may cause poor results toward the minority labeled class by a DL model. DL is the mainstream approach to process ocular images due to its capability of extracting features. However, the black box representation in terms of interpretability, adaptation, and reasoning-ability may lead to poor results~\cite{alaa2019demystifying}. Regarding the aforementioned limitations, this survey discusses the basic principle of NeSyL, and proposes its potential applications in ophthalmic diseases, diagnosis and prediction, and circumventing the limitation of DNNs.
	\begin{table*}[h!]
		\caption{The summary of AI applications in classification, segmentation and prediction in recent years.\\ MSCNN: Multiscale sequential convolutional neural networks, RPNCN: Region proposal network and cascaded network, ALMF: Adversarial learning with multiscale features, age-related eye disease study (AREDS), Indian diabetic retinopathy image dataset (IDRiD) } 
		\label{tab:AI_APP}
		\setlength\tabcolsep{1pt} % let LaTeX compute intercolumn whitespace
		%\footnotesize		
		\smallskip 
		\centering
		\renewcommand{\arraystretch}{1.2}
		\begin{tabular}{p{2.3cm}|p{1.6cm}|p{1.7cm}|p{2cm}|p{5cm}|p{5cm}} \hline 
	{Reference}&{Task}&{Disease}&{Method}&{Dataset}&{Results}\\ \hline
		
	{Bajwa et al.~\cite{bajwa2019two}}&{Classification}&{Glaucoma} &{RCNN \& DCNN}&{ORIGA}&{AUC: 0.868}\\ \hline
		
	{Peng et al.\cite{peng2019deepseenet}} &Classification&AMD&	De.pdfeeNet	&AREDS	&Patient-based classification accuracy = 0.671, retinal specialists of 0.599\\ \hline
			
	{Zhang et al.\cite{zhang2020attention}}&Classification&Cataract&	Ensemble attention model&Private dataset (1,877 ultrasound images of normal eyes and 1,896 of cataract eyes)&AUC: 0.975\\ \hline
			
	{Wan et al.\cite{wan2018deep}}&Classification&DR&CNNs&Kaggle&AUC:0.97\\ \hline	
				
	{Prahs, P. et al. \cite{prahs2018oct}}&Classification&AMD&Inception&The electronic intervention records from Django Project&AUC: 0.988; specificity: 0.941; sensitivity: 0.942\\ \hline	
		
	\multirow{3}{*}{Yan et al. \cite{yan2018joint}}&\multirow{3}{*}{Segmentation}&	\multirow{3}{*}{Retinal Vessel}&\multirow{3}{*}{ U-Net like AE}&DRIVE&AUC: 0.9752 \\ 
	&&&&STARE&AUC:0.9801\\ 
	&&&&CHASE\_DB1&AUC:0.9781\\	\hline
	
	\multirow{2}{*}{{Akram et al. \cite{akram2019adversarial}}}&\multirow{2}{*}{Segmentation}&	\multirow{2}{*}{Retinal Vessel}&\multirow{2}{*}{ ALMF}&DRIVE&AUC: 0.9890; F1-score: 0.8003; Specificity: 0.7851; Accuracy: 0.9659 \\ 
	&&&&STARE&AUC: 0.9860; F1 score: 0.7710; Specificity: 0.7634; Accuracy: 0.9812.\\ 	\hline

	{Fu et al.~\cite{fu2018joint}}&segmentation&Glaucoma&M-Net&ORIGA&Edisc: 0.083; Adisc: 0.972; Ecup: 0.256; Acup: 0.914; Erim: 0.265; Arim: 0.921; E:0.078\\ \hline	
	
	{Al-Bander et al~\cite{gu2018deepdisc}}&\multirow{2}{*}{Segmentation}& \multirow{2}{*}{Retinal Vessel}&\multirow{2}{*}{MSCNN}&MESSIDOR &OD center detection Accuracy :97\%; Fovea center detection Accuracy:96.6\% \\ 
	&&&&Kaggle&OD center detection Accuracy: 96.7\%; Fovea center detection Accuracy:95.6\%\\ \hline	
	
	{Huang et al. \cite{huang2020efficient} }&\multirow{2}{*}{Segmentation}&\multirow{2}{*}{Retinal Vessel}&\multirow{2}{*}{RPNCN}&IDRiD&OD detection accuracy: 100\%; Fovea localization accuracy:99.03\% \\
	&&&&Messidor &OD detection accuracy:100\%; Fovea localization accuracy:99.25\%.\\ 	\hline	
	
	{Yang X et al. \cite{yang2020prediction}}&{Prediction}&{Myopia}&{Support Vector Machine (SVM)}&{3112 students in grade 1 to grade 6, including their individual activity, eye condition, parental heredity and so on.}&{10-fold cross-validation accuracy: 0.92; precision: 0.95; sensitivity: 0.94; f1-score: 0.94; AUC: 0.98
	}\\ \hline
	
	{Varadarajan et al. \cite{varadarajan2018deep}}&Prediction&Refractive error& DNN with attention&The UK Biobank and AREDS&Mean absolute error (MAE): 0.56 diopters on the UK Biobank and 0.91 diopters on the AREDS\\ \hline

	{Apreutesei et al. \cite{apreutesei2018predictions}}&Prediction&	Glaucoma&	Jordan Elman networks (JEN) 	&52 patients (101 eyes) with glaucoma and diabetes&Accuracy: 0.95\\ \hline

	{Jiang et al. \cite{jiang2018predicting}}&Prediction& Posterior capsular opacification&	Temporal sequence network (TempSeq-Net)& 6,090 slit-lamp images from the Zhongshan Ophthalmic Center of Sun Yat-sen University	&Accuracy: 0.922; sensitivity: 0.886; specificity: 0.943; AUC: 0.972\\ \hline
 \end{tabular}
		\justify  
	\end{table*}
	
	\section{Principle of Neuro-Symbolic AI}
	\subsection{Symbolic AI versus Connectionist AI}
\begin{figure}[!h]
			\centering
			{\includegraphics[width=0.5\textwidth]{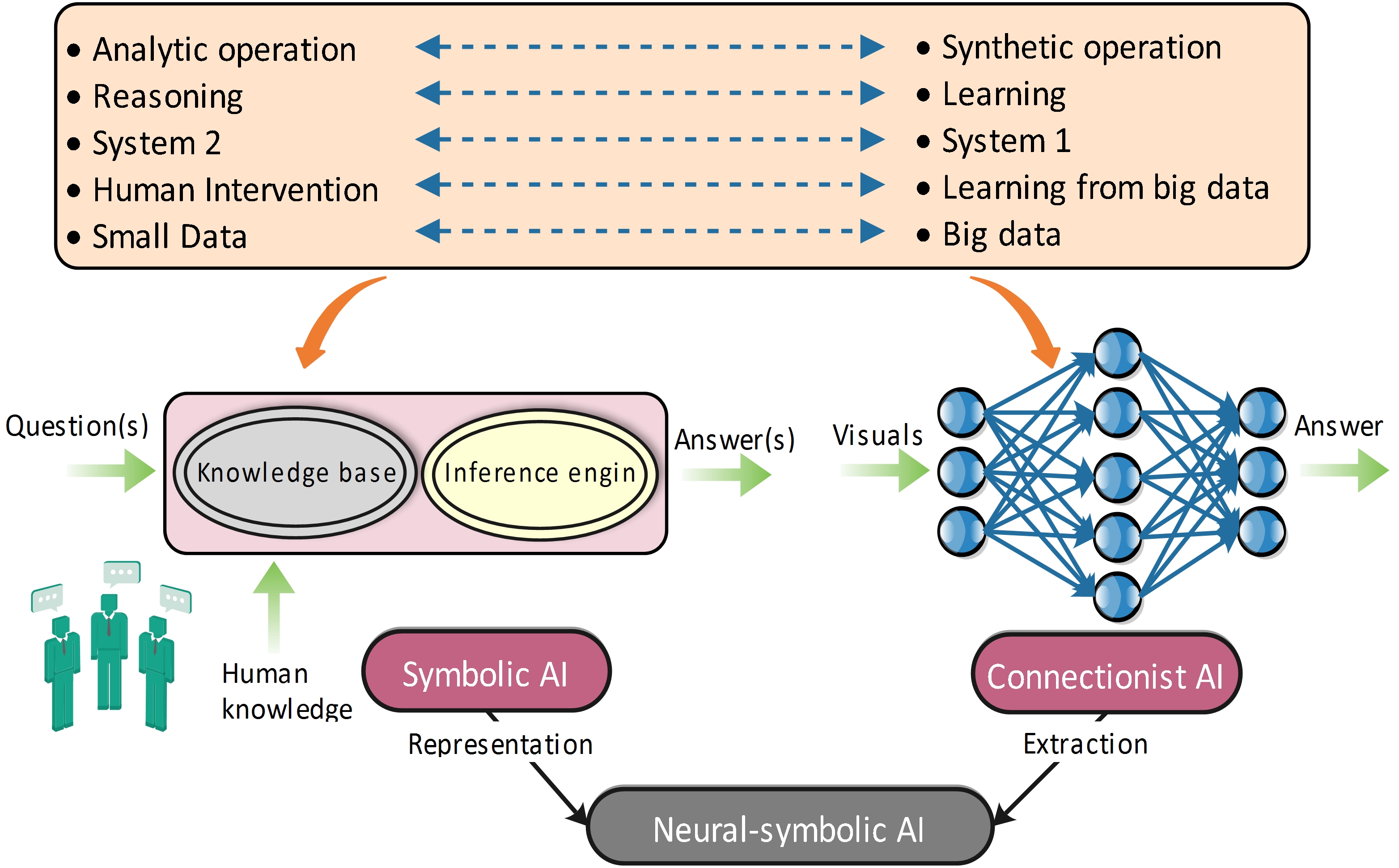}
			}
			\caption{Symbolic AI vs connectionist AI.}
			\label{fig:symbolic_AI_connectionis}
		\end{figure}

	For many years, there have been two main schools of thought in AI fields, namely symbolic AI (sAI) and connectionist AI (cAI). Hubert and Stuart Dreyfus \cite{dreyfus1991making} have defined the correlation between sAI and cAI by phrasing as: making a mind versus modeling the brain. According to human cognitive ability, the creation of the mind (sAI) usually starts at a higher cognitive level through symbolic manipulation and language processing. On the other hand, the creation of the brain (cAI) starts at the lowest cognitive level by developing neurons and their interconnections.  Moreover, Kahneman \cite{kahneman2011thinking} proposed a concept for a binary architecture (called System-1 and System-2) to explain the results of processing speed differences in humans on an assortment of mental tasks. Similarly, the cAI could be classified as System-1, while sAI is considered to belong to System-2.  Figure~\ref{fig:symbolic_AI_connectionis} shows the main differences between sAI and cAI.
	\subsubsection{Symbolic AI}
	The sAI has a list of aliases, including classic AI, rule-based AI, and good old-fashioned AI (GOFA) \cite{haugeland1989artificial}. From the mid-1950s until the middle 1990s, much of the AI research focused on sAI, which depends on embedding human knowledge and behavioral rules into computer codes \cite{quinlan1994comparing}.
	
	The sAI assumes symbols are the basic units of human intelligence. Similarly, the human cognitive processes are a series of explicit inferences about symbolic representations. From a cognitive point of view, sAI corresponds to analytical operations, which are mostly related to high-level, decomposable, conscious, and reasoning tasks. In analytical operations, there exists a conscious thought process and conscious decisions using explicit high-level information. The knowledge involved in symbolic and propositional knowledge is concerned with language. The sAI and analytical operations are found to share many features such as sequential decomposability and propositional knowledge. The application of sAI processes strings of characters, representing real-world entities or concepts through symbols. All the symbols can be arranged hierarchically or via lists and networks, which dictate the algorithm association among symbols \cite{cost1993weighted}.
	
	The sAI system often refers to logical inference, deductive reasoning, and some search algorithms that aim to solve the constraints of a specified model \cite{sun2013connectionist}. An expert system belongs to this system, which is based on human-created knowledge and generally contains if/then pairing statements instructing algorithms behaviors. An inference engine model refers to the knowledge-base and selects the rules corresponding to a given symbol or symbol sets. The sAI has been widely used in tasks that have clear-cut rules and goals \cite{ortega2021symbolic,gocev2020supporting}. However, the algorithmic complexity of discrete reasoning is nondeterministic polynomial time (NP) hard in such algorithms requiring substantial hand-tuning, which made them difficult to address complex problems. The sAI faces difficulty in modeling uncertain and ambiguous knowledge. The attempt of solving real-world tasks and super-massive spaces is a challenging problem.
	\subsubsection{Connectionist AI}
	The cAI receives its name from the typical network topology employed in this category. It is the dominant paradigm in the AI field because of the technological successes of connectionism in the presence of large data sets. The cAI models the processes based on how the human brain works via its interconnected neurons, which usually applies a perceptron to represent a single neuron. From a cognitive point of view, it corresponds to synthetic operations, and they are mostly concerned with low-level, unconscious, non-decomposable, and perceptual tasks. Vision and speech recognitions are the typical tasks that involve the knowledge limitations to express verbally and in a non-propositional way. However, the effective use of semantic knowledge can improve CNN performance. 
	
	Typically, cAI systems learn associations from large-scale data with no or little prior knowledge. On the contrary, sAI can be built via hand-coded programs by humans.The cAI is composed of all kinds of neural networks, such as CNNs, DNNs, and graph neural network (GNN). The cAI offers good performance in pattern recognition and generalization. However, it often cannot explain how it arrived at a solution, which is highly needed for safety-critical fields such as self-driving cars and medicine. It also requires a wealth of data to ensure high training and testing accuracies, which is unrealistic for some tasks.
	\subsubsection{Symbolic and Connectionist AI}
	Many scientists reach an agreement that information at different levels of abstraction differs in structure and makeup. They assume that lower levels of abstraction are subsymbolic and higher levels are symbolic. Hugo Latapie \cite{latapie2021neurosymbolic} proposed that cognition is not dualistic in this way and that knowledge at any level of abstraction involves neural-symbolic information, in other words, both symbolic and subsymbolic information are contained in data at both levels. This implies the realization of neural-symbolic models as natural mimic machine for information processing, which could cumulate connectionist (cAI) and symbolic (sAI) capabilities.
	
	Neural-symbolic AI or NeSyL could extract features from data using connectionist approaches and then manipulate these features via symbolic approaches. Neural-symbolic AI has proven to converge quickly with only the 10$^{th}$ part of the training data. Scientists believe that neural-symbolic AI will empower AI with the ability to learn and reason while performing a wide variety of tasks without extensive training. Symbolic component regularizes the neural learning, while neural components help model scaling and guide discrete decisions. 
	\subsection{Integration Methods: Hybrid Versus Unified}
	Various methods are used to integrate neural and symbolic AI. Generally, different tasks and researches are suitable for using different integration methods. NeSyL is more difficult to be implemented in parallel computation than traditional neural models due to complex control flow and low-operational-intensity operations. The neural learning of computation dominates the symbolic reasoning when they are clearly separable. In 1995, Yannic et al. made an effort to lay the cognitive foundation for neurosymbolic integration, and described different strategies employed up to that time \cite{lallement1995neurosymbolic}. The study \cite{lallement1997cognitive} identified two main methods of constructing neural symbolic integration models: hybrid and unified.
	\subsubsection{Hybrid Methods}
	Hybrid methods combine independent submodules, including several neural and symbolic submodules. Each submodule has its own role and performs different functions. Neural submodules are mainly responsible for perceptual tasks, while symbolic submodules are responsible for inference analysis. Translational and functional hybrids are the two hybrid approaches summarized by Hilario et al. \cite{hilario1997overview}, in which the functional hybrids have been further classified into chain processing, sub-processing, meta-processing, and coprocessing. Garcez et al. presented their study of hybrid approaches in \cite{avila1999connectionist,garcez2007connectionist,garcez2006connectionist}, as early efforts for neurosymbolic integration. In recent years, a number of studies \cite{mao2019neuro,yi2019clevrer,han2019visual,chen2021grounding} on neuro-symbolic have utilized hybrid approaches. The submodels in hybrid systems can be constructed separately, in which the hybrid methods can directly benefit from advances in neural networks and symbolics. However, hybrid systems may not reflect biological reality \cite{hilario1995neurosymbolic}. Gary Marcus claims that Hybrids are often effective~\cite{marcus2020next}.
	\subsubsection{Unified Methods}
	Unified methods utilize both connectionism and symbolic ability by adding another paradigm to neural or symbolic approaches. The unified methods are further divided into two cases. The first case integrates symbols into the neural network and the second constructs neural architectures for symbolic processing. A unified approach can also be viewed as an end-to-end approach with exception of a submodel. However, a unified model can be used as submodel in a hybrid system. As Garcez et al. proposed logic belonging to the first case can be thought of as the language for compiling neural networks~\cite{garcez2020neurosymbolic}. The unified architecture of the brain should be equipped with biological plausibility and may lead to discoveries. In addition, unified systems have the advantages of computational efficiency and biological fidelity in massively parallel processing. However, unified approaches are less scalable \cite{hilario1995neurosymbolic}.
	
	\subsection{NeSyL Components}
	In this section, we describe the components of a neuro-symbolic system, which include attention mechanisms, knowledge graphs, and representation methods. They can collaborate to achieve better results.
	\subsubsection{Attention Mechanism}
	Bahdanau \cite{bahdanau2014neural} introduced an attention mechanism in machine translation to solve the fixed-length encoding problem successfully and retroactively. The attention mechanism was computed separately for the alignment score, weight, and context vector. The alignment model was designed as a feed-forward neural network, which was trained with other components. The decoder structure was designed to pay extra attention to a certain part of the data set. Thus, the alignment mechanism increases the accuracy of prediction facilitated by the attention mechanism to focus only on the relevant contents. 
	
	It is important to remove irrelevant terms from the symbolic reasoning process where the excessive number of logical combinations makes computation challenging. Symbolic distillation and relaxation of symbolic programs into neural networks convert combinatorial search over symbolic code to scalable gradient-based techniques. Distillation emerges as an alternative solution for learning symbolic and neurosymbolic programs. Latapie \cite{latapie2021neurosymbolic} applied an attentional mechanism to neuro-symbolic reasoning to filter huge combinations of data and realize parallel calculations of different combinations.
	\subsubsection{Knowledge Graphs}
	The graph is the associated link of all elements or vertices. The knowledge base and the reasoning machine jointly form a knowledge system. The graph can serve as a knowledge base to store knowledge and perform reasoning~\cite{fensel2020introduction}. Graphs link these entities by mimicking the connections of neurons in the brain. Neuro-symbolic systems also have this capability, benefiting from knowledge graphs as a bionic technology. Graphs enable AI to reason and learn based on common sense, prior knowledge, and statistics. For a graph-based AI, new knowledge will be integrated into the larger graph via logics. Such models can learn new knowledge continuously even after completing training which saves computations in large systems~\cite{chaudhri2022knowledge}. In addition, the combination of the attention mechanism and the graph is designed to compute independent relationships among neighboring nodes' weights of a graph~\cite{wang2020multi}. 
	\subsubsection{Representation Methods}
	Knowledge representation is considered to be one of the core concepts of AI, which combines objects, logical representation, and semantic networks~\cite{markman2013knowledge}. Logic represents a rule without ambiguity such as propositional logic, production rules, frames, and predictive logic. Programming languages are designed based on logical representations which allow the logical deduction. Ambiguity in natural language may lead to unclear representations and it is difficult to correspond to logical representations \cite{keselj2009speech}. The knowledge can also be represented via semantic network to classify objects and represent interrelationships through wires to bring ease to network expansion. The gap between semantic networks and knowledge graphs need to be filled for a better knowledge representation \cite{chaudhri2022knowledge}.
	
	On the other hand, knowledge representation has two routes, localist and distributed representation~\cite{garcez2020neurosymbolic}. The localist representation has advantages of interpretation and high computational efficiency on simple models. However, it has obvious disadvantages of loss relevant features to be encoded as one-heat codes. Distributed representation is more conducive to generate dense data, which is favorable to improve the computational efficiency of complex features. Most importantly, it preserves the correlation between features~\cite{liu2016knowledge}. Distributed representation is applied to compute similarity and supplement the knowledge base distributed representation \cite{liu2016knowledge}. Distributed representation has been widely used in natural languages and entities \cite{klementiev2012inducing,zhao2015representation}. The structured embedding (SE) \cite{garcez2020neurosymbolic}, single layer model (SLM) \cite{socher2013reasoning}, semantic matching energy (SME) \cite{bergen2021jointly,bordes2014semantic}, latent factor model (LFM) \cite{jenatton2012latent,sutskever2009modelling}, and neural tensor network (NTN) \cite{socher2013reasoning} have been developed to implement the knowledge representation. Compared with knowledge graph, knowledge representation can save computational cost and improve accuracy \cite{liu2016knowledge}. The knowledge representation potential will be further unleashed after combining with neural networks.
	
	\subsubsection{Decision Rules}
	The decision rules are an indispensable component in the incorporation of knowledge representation. Decision rules are the If-The statements with constraints that have similarities to natural language and is suitable for interpretation. To avoid overlap and rule missing, decision lists and decision sets are proposed to solve the combination problem \cite{molnar2020interpretable}. The rules in the decision set are given different weights based on properties such as accuracy. Algorithmic processing of the data can produce new rules. The OneR algorithm \cite{holte1993very} uses the entire dataset to learn and obtain multiple rules according to the produced error. In addition, suitable intervals are used to divide the continuous values to extract discrete features. However, for a dataset containing multiple rules, OneR is not applicable. Cohen [65] proposed RIPPER as an improvement on sequential covering, which applied rule pruning for optimization \cite{molnar2020interpretable}. Bayesian Rule Lists are another way of rule learning. The first step of this approach applies pre-mining of frequent patterns using frequent pattern (FP) growth or Apriori~\cite{borgelt2005implementation}. There are other algorithms used to generate decision rules~\cite{zabinski2020decision}. Decision rules have been used together with deep learning \cite{okajima2019deep}. Decision rules are considered suitable for interpretation and fast prediction but may lack regression problems~\cite{molnar2020interpretable}. 
	
\begin{figure*}[h!]
		\centering
		\includegraphics[width=1\linewidth]{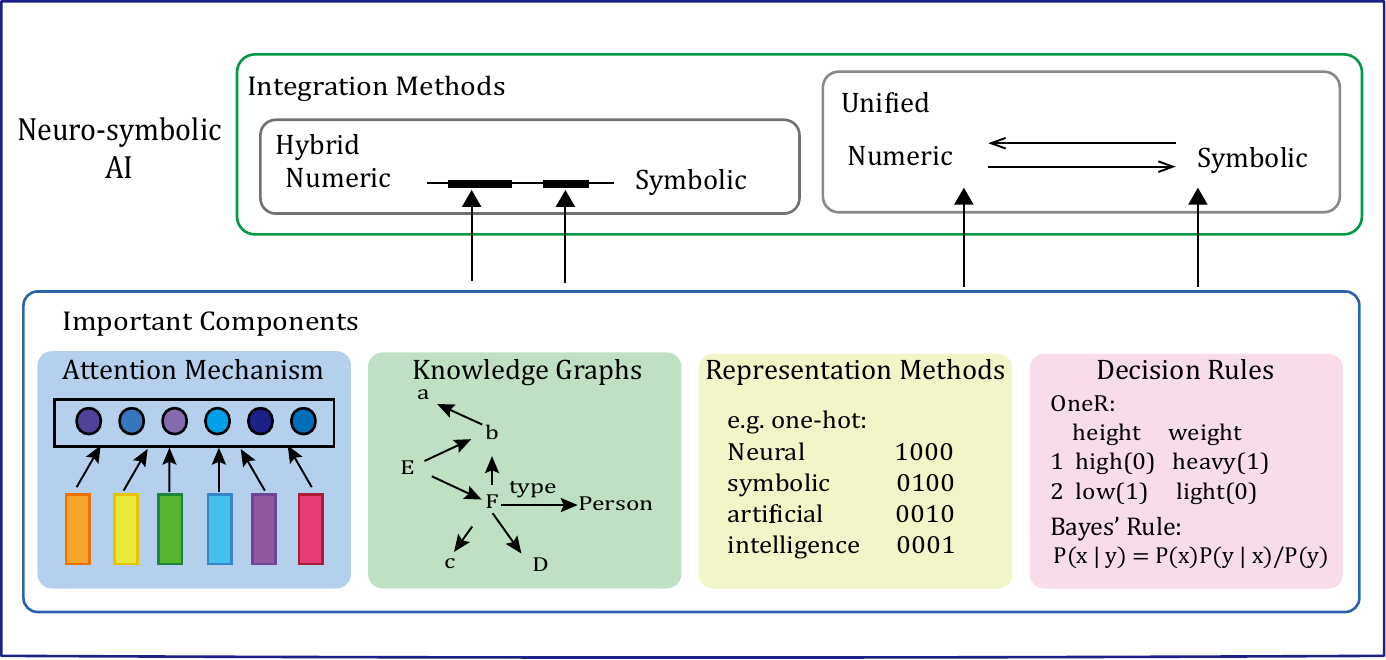}
		\caption{Association between integrated maethods and important components of Neuro-symbolic AI.}
		\label{fig:NSAI-components-integration}
	\end{figure*}

	NySyL Components are closely related to the integration methods of Neuro-symbolic AI (Figure~\ref{fig:NSAI-components-integration}). Components such as attention mechanisms and knowledge graphs can be used in each submodule in hybrid integration methods. In the unified methods,  such components can be added to another paradigm to form single neural or symbolic arithmetic. 
	
	\subsection{Task Adaptive NeSyL Structure}
	Various categorizations have emerged trying to classify neuro-symbolic structures. In this section, we present the most widely accepted scenarios followed by some leading examples in the respective field. The first approach to neuro-symbolic taxonomy is presented in~\cite{chaudhuri2021neurosymbolic} where 5 different structures are illustrated in Tabl~\ref{tab:Five_structure}.

	\begin{table}[h!]
		\caption{Overview of 5 structures presented in the first approach to neuro-symbolic taxonomy} 
		\label{tab:Five_structure}
		\setlength\tabcolsep{2pt} % let LaTeX compute intercolumn whitespace
		%\footnotesize		
		\smallskip 
		\centering
		\renewcommand{\arraystretch}{1.2}
	\begin{tabular}{p{2.1cm}|p{6.4cm}}\hline 
		{Structure}&{Description}\\ \hline
		{Symbolic-after-neural pipeline}&{The input is first processed by neural components and then fed to either logic or functional symbolic codes. The Houdini  is an example of such a concept, which consists of a symbolic program synthesizer that searches over a library of functions and a neural module to train these functions~\cite{valkov2018houdini}.}\\ \hline
	Neural-after-symbolic pipeline& The input first passes via a set of synthesized program components followed by a neural network.\\ \hline
	Neural module networks& Several neural modules appear as components to provide high-level functionality (network layout predictor, modular meta-learning).\\ \hline
	Algebraic compositions& Neural and symbolic programs run independently and then results are combined with algebraic operators such as \textbf{co}ntrol \textbf{re}gularized \textbf{r}einforcement \textbf{l}earning (CORE-RL) and imitation-\textbf{pro}jected \textbf{p}rogrammatic r\textbf{e}inforcement \textbf{l}earning (PROPEL).\\ \hline
	Neurally accelerated symbolic programs& Machine learning is used to guide the execution of a program without affecting its functionality and improved its efficiency (safe learned controllers).\\ \hline
		\end{tabular}
		\justify  
	\end{table}
		
\begin{figure}[h!]
	\centering
	{\includegraphics[width=0.45\textwidth]{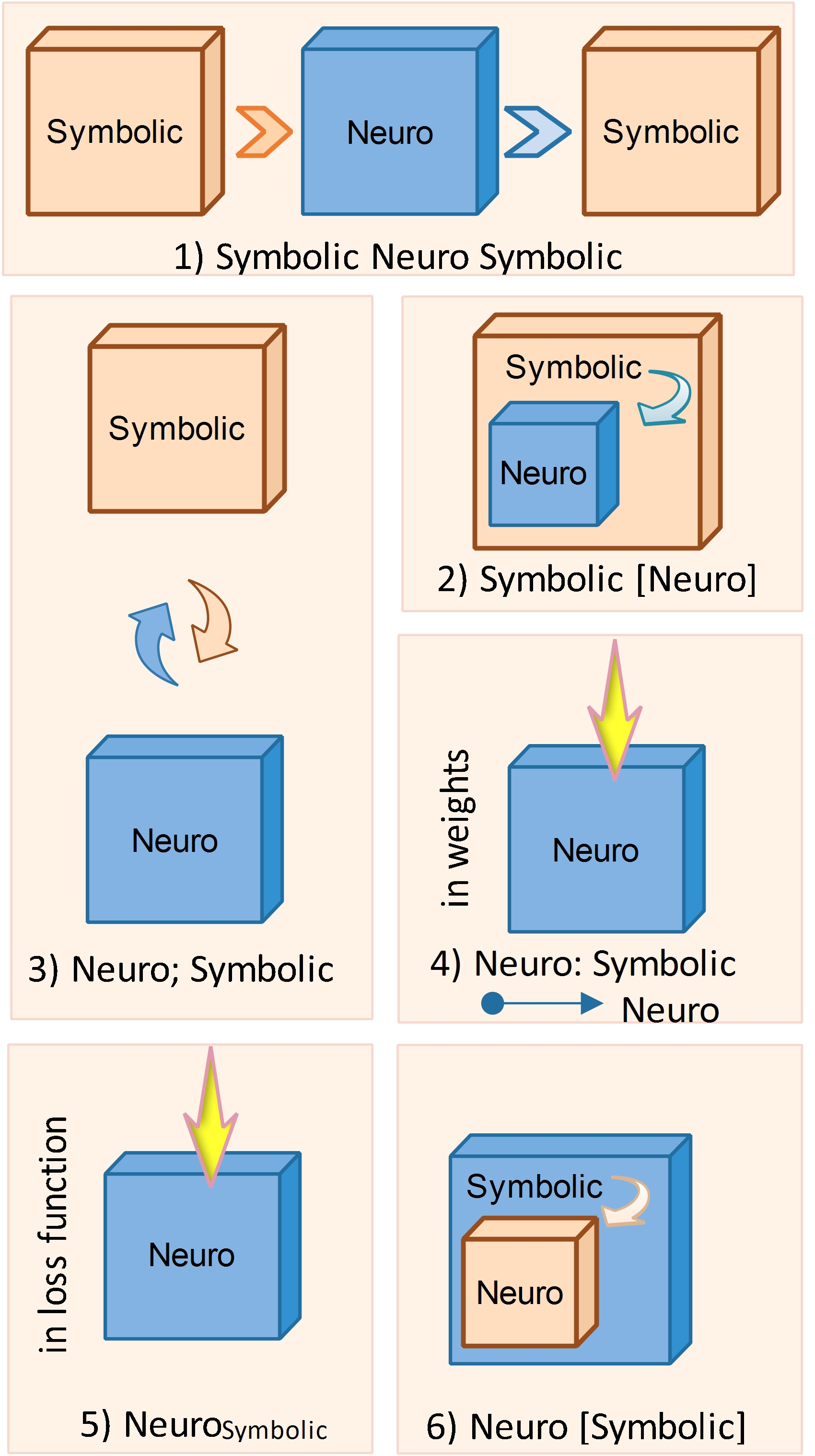}
	}
	\caption{Neurosymbolic taxonomy by Professor Kautz.}
	\label{fig:NSL-toxonomy-Kautz}
	\end{figure}

	Another approach to taxonomy has common characteristics with the illustrated one (Table~\ref{tab:Five_structure}) and is largely accepted by the literature~\cite{garcez2020neurosymbolic, sarker2021neuro, lamb2020graph,hamilton2022neuro} (Figure~\ref{fig:NSL-toxonomy-Kautz}). Pointer networks or Graph Neural Networks (GNNs) may be deployed to solve graph problems with attention mechanisms (Figure~\ref{fig:NSL-toxonomy-Kautz}).
	We provide an overview of the 6 structures in Table~\ref{tab:six_structure}.
	\begin{table}[!t]
	\caption{Overview of 6 structures presented by Professor Kautz} 
	\label{tab:six_structure}
	\setlength\tabcolsep{2pt} % let LaTeX compute intercolumn whitespace
	%\footnotesize		
	\smallskip 
	\centering
	\renewcommand{\arraystretch}{1.2}
	\begin{tabular}{p{2cm}|p{6.5cm}}\hline 
		{Structure}&{Description}\\ \hline
		{Symbolic Neuro Symbolic}&{All symbols are converted to vector embeddings followed by the processing of neural models and produces symbol as the standard deep learning procedure.}  \\	\hline
	{Symbolic [Neuro]}&{Refers to symbolic solvers, such as the Monte Carlo tree search, that use neural models as subroutines, e.g., neural state estimator.}\\ \hline
	
	{Neuro; Symbolic}&{Neural and symbolic systems are assigned different tasks in the parallel pipeline and communicate the extracted information to improve the individual or collective systems performance (e.g. neuro-symbolic concept learner,  deepProbLog).}\\ \hline
	
	{Neuro:Symbolic $\rightarrow$ Neuro}&{The symbolic knowledge is compiled into a neural network model. Various forms of symbolic knowledge is embedded in the neural network by assigning its initial weights.}\\ \hline
	
	{NeuroSymbolic}&{A symbolic logic rules transformation regularize the network’s loss function. Such rich transformation and logic embeddings are present in Logic Tensor Networks (LTNs) and Tensor Product Representations.}\\ \hline
	
	{Neuro [Symbolic]}&{The neural network model performs combinatorial symbolic reasoning. This is achieved either by learning the relations between the symbols or by paying attention to selected symbols at a certain point (e.g. GNNs).}\\ \hline
	\end{tabular}
	\justify  
\end{table}

	\subsection{Proposed Large Frameworks}
	This section presents frameworks that have attracted attention as building components in NeSyL. Among these frameworks, DeepProbLog is a probabilistic logic programming language that uses neural predicates to accomplish deep learning~\cite{manhaeve2018deepproblog}. DeepProbLog stems from ProbLog, while maintaining all its functionalities~\cite{manhaeve2019deepproblog}. Moreover, the pure neural network, logical and probabilistic methods are presented to allow varieties of framework combinations~\cite{hitzler2022neuro}.
	
	The Neuro-symbolic concept learner is a model that learns visual concepts by parsing sentences without explicit supervision~\cite{mao2019neuro}. The induction approach extracts novel concepts from noisy perceptual experience as by mimicking childrens' ability to label objects without being explicitly taught. In such concept, first an object-based scene representation is built together with translated sentences in parallel into executable symbolic programs. These two modules are integrated by a neurosymbolic reasoning module. The structured representation following a transfer learning might be applicable to different tasks and the replacement of attributes with placeholders allows new attributes to be learnt~\cite{berlot2021neuro}.
	
	Neuro symbolic forward reasoner (NSFR) is a framework for reasoning tasks using first-order logic and converts raw input into object-centric representations. For differential forward-chaining, weighted rules follow to generat probabilistic logical atoms~\cite{shindo2021neuro}. The whole process is divided into three components: I) the object-centric perception module, ii) the facts’ converter, and iii) the differentiable reasoning module. In this way, a new set of atoms can be deduced from given facts.
	
	Commonsense transformers (COMET) is a generative model based on commonsense knowledge that learns to generate commonsense descriptions in natural language resulting in the construction of a knowledge base(KB)~\cite{bosselut2019comet}. COMET extracts KB structure, relations, and representations among nodes and edges to seed the knowledge to a graph.
	
	Bidirectional encoder representations from transformers (BERT) is a language representation model, which is designed to pre-train deep bidirectional (both left and right context) representations from unlabeled text, and then fine tunes with labeled data~\cite{kenton2019bert}. This makes BERT appropriate for various tasks. During training and fine-tuning the same architecture is used except for the last layer. Another important feature is the questions answering representation into token sequence which makes it appropriate for visual question answering (VQA) applications.
	
	Logic tensor networks (LTNs) is a statistical relational learning (SRL) framework that applies real logic constants and interprets function symbols as real-valued functions and fuzzy logic relation
	
	~\cite{donadello2017logic}. 
	
	Therefore, the soft and hard constraints on real-valued data can be presented in first-order logic to improve learning constraints~\cite{serafini2016logic}.
	
	Graph neural networks (GNNs) generally learn an embedding that maps nodes and edges into a continuous vector space with limitations on the number of vertices in the input graph~\cite{lamb2020graph}. This approach has been enriched with attention mechanisms. For instance, a multi-hop attention graph neural network (MAGNA) first computes the attention scores on all edges and then the attention diffusion module assigns attention scores between pairs of nodes that are not directly connected~\cite{wang2020multi}.

	\section{Revisiting CNN through NeSyL}
	\subsection{Ocular Applications of Neural Network}	The most widely used image recognition methods such as CNN \cite{babu2022efficient, nusinovici2022retinal}, GCN \cite{song2021graph} can be used to identify, diagnose and infer from ocular images. For diagnosing glaucoma, a double tier deep convolutional neural network was used with convincing accuracy rate of 92.64\%~\cite{babu2022efficient}. Nusinovici et al. \cite{nusinovici2022retinal} used a deep learning approach to predict biological age based on retinal photographs. Upadhyay et al \cite{upadhyay2022coherent} used coherent convolutional neural networks to detect retinal disease on optical coherence tomography images. Neural network-based methods have the advantages of availability, robustness, and accuracy.
	
	Neural Network and Deep Learning are very popular these days for their powerful magic in solving varieties of problems; however, uninterpretability and explainability like challenges still need to be addressed. Although by complex network architecture and computational approaches, DL models can produce a viable solution to many challenging problems; however, the latent space under the jillion numbers in the latent space is unexplainable to  humans. It is inevitable to exploit the principles behind the results produced in the legal domain. Good old fashioned AI (GOFAI), also known as symbolic AI, is an interpretable and generalized concept equipped with reasoning capability. However, the GOFAI needs to manually establish the knowledge dataset in the representation form of symbols. Thus, it is very natural to bring these two concepts together into NeSyL. 
	\subsection{Potential Superiority of NeSyL over Classical Deep Learning}
	NeSyL can be applied to many fields. Compared with the classical DNN, NeSyL has the advantages of lower cost, better generalization and robustness, interpretation and explainablity, efficient training, and mimicking human cognitive ability.
	
	Several efforts \cite{akbari2021does} have already been made to demonstrate that the loss function matters for the generalization abilities of DNN. Some researchers have used the symbolic reasoning knowledge to optimize the learning of the DNN and derive a better loss function \cite{xu2018semantic}. Therefore, applying a NeSyL before the DNN to choose the loss function may help improve the performance of the entire system. A GNN may not compute  important graph properties without additional information~\cite{garg2020generalization}. Common methods like optimizing the structure or adding additional message such as labeling can be a huge cost in DL. CNN model can be pruned by specializing the polynomial to train the regression more effectively~\cite{wang2021accelerate}, which is similar to synthesis in the NeSyL.  In comparison, NeSyL has more advantages for its low cost in labeling.   Researches have pointed out that both the depth and the width of a network matter in the deep learning, which makes it harder to balance the time and resource consumption~\cite{nguyen2018optimization}. Compared with CNN, the NeSyL needs few resources and can achieve satisfying performance in many situations.
	
	The study [91] proposed that DNNs are vulnerable to adversarial perturbations which can be solved by considering domain knowledge such as road signs in autonomous vehicles. NeSyL can produce a robust system by applying the knowledge enhancement strategy. A system generalization is also a well-worth paying attention sector; however, both generalization and domain adaptation is hard to jointly improve~\cite{zhao2019learning}. Henceforth, deep learning has its limits in domain adaptation and representational learning, where the NeSyL can perform much better. There is also a problem of overfitting in deep learning networks which weaken the robustness to adversarial attacks~\cite{rice2020overfitting}. It is hard to explain the observed overfitting. NeSyL maybe helpful in the researching fields for its advantages such as domain adaptation, interpretations, and auto reasoning.
	
\begin{figure}[h!]
		\centering
		\includegraphics[width=1\linewidth]{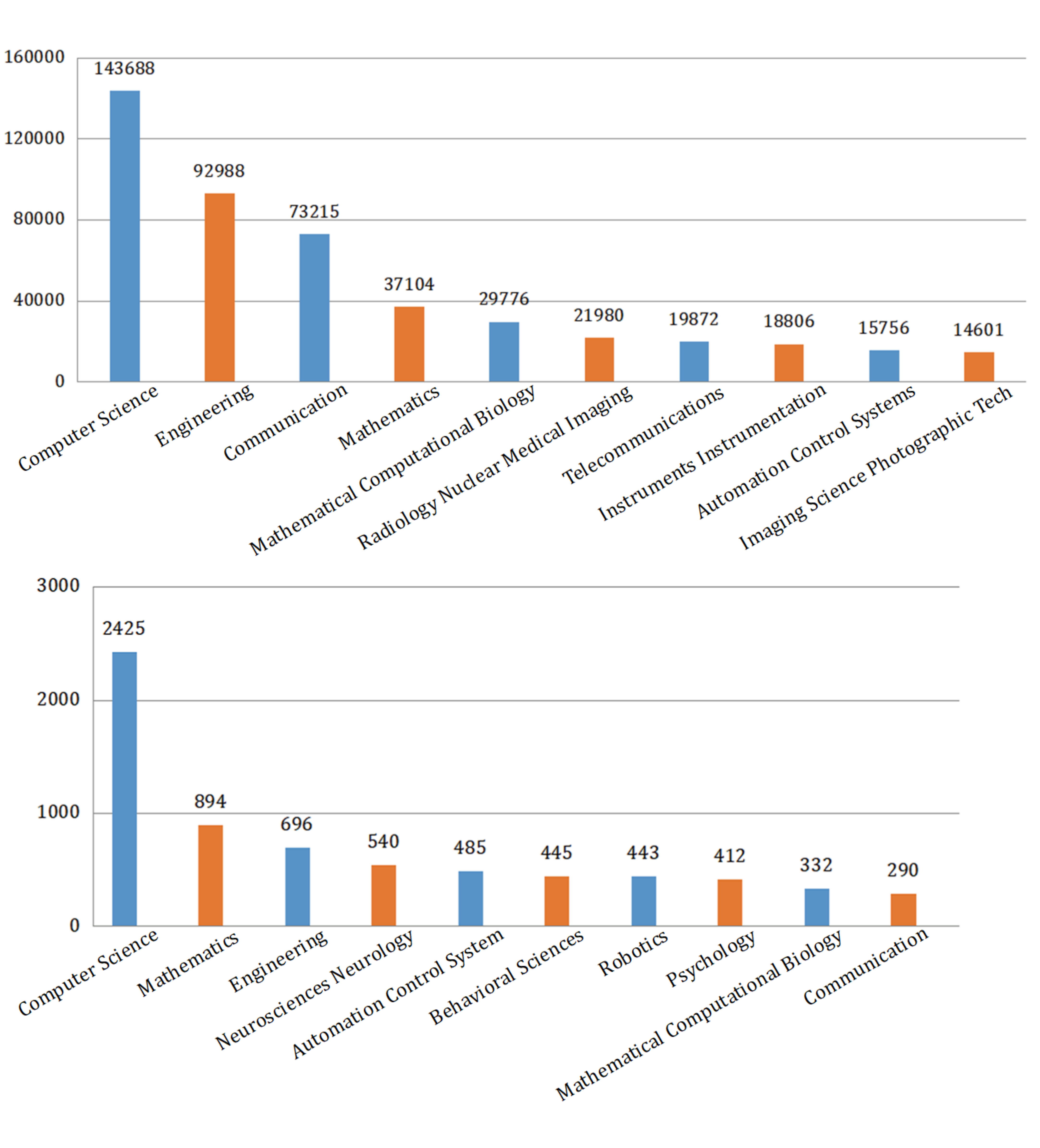}
		\caption{Comparative depiction of CNN and NeSyL approaches in distinct fields of life. Top figure shows the top-10 research field in the neural symbolic field and the bottom figure shows the applications of CNN in the top-10 research fields.}
		\label{fig:Top10-Research-Field-NSL}
	\end{figure}

	Moreover, large scale pretraining networks have been leveraged to do the neural symbolic regression and satisfying performance in re-discovering physical equations~\cite{biggio2021neural}. Equivariant network, a promising fusion of NeSyL and image processing is a capable model to improve the performance and efficiency of the symmetrical neural network in computer vision and medical imaging which can be extended to wide applications~\cite{sen2021neuro}. An overlap can be observed for the top 10 research fields of CNN and the NeSyL, such as Computer Science, Engineering and Mathematical Computational Biology (Figure~\ref{fig:Top10-Research-Field-NSL}). However, efforts have been made on  CNN while the NeSyL received relatively less attention. Considering the advantages over traditional networks, more attention is needed for in-depth research in NeSyL.

	\subsection{Image Processing via NeSyL}
    The CNN models contributed tremendously in image processing during the last decade specifically in medical imaging and ocular diseases diagnosis. On the other hand, NeSyL integrates expert knowledge and adds explainability to ensure transparent decision system to humans. A transparent system in clinical domain becomes a requirement; thus, the research potential and progress in the field of NeSyL is replacing CNN in ocular images. An explainable and interpretable NeSyL method was utilized for diabetic retinopathy classification. They used a U-Net segmentation network to extract human-readable symbolic representation proceeded to a fully connected network for decision making purposes. 
        
    The NeSyL model obtained comparable performance to state-of-the-art methods on Indian diabetic retinopathy image dataset (IDRiD) featured with achieving interpretability and explainability. To combine the representative power of deep learning and the interpretable capabilities of symbols, Chowdhury et al. \cite{chowdhury2021emergent} demonstrated an emergent language based classification framework that involved a sender, generator, and a receiver for medical image classification. The sender extracts feature representations and feeds into a symbol generator to generate symbols. The receiver accepts generated symbols to perform the classification. Han et al. \cite{han2021unifying} provided a novel method for generating spinal medical report automatically. They embedded a symbolic graph based reasoning module into an adversarial network for semantic segmentation of spinal structures. Based on these outputs from neural learning, the framework conducted causal effect analysis to detect abnormalities through symbolic-logical reasoning and generated spinal medical reports. The pool of concepts influences the semantic information contained in the system and logical reasoning.  
	
	To tackle this problem, Diaz-Rodriguez et al. \cite{diaz2022explainable} leveraged symbolic representations and presented the explainable NeSyL (X-NeSyL) methodology to fuse deep learning representations with expert domain knowledge which offers faced images classification and architectural element annotations. Alirezaie et al. \cite{alirezaie2019semantic} designed a semantic referee which extracts features from errors produced by a deep convolutional network as additional image channels to assist miss classification correction. Although, neural symbolic methods perform impressively in visual question answering on synthetic images, they suffer from the long-tail distribution of visual concepts and unequal reasoning st.pdf which are two challenges in real images. Li et al. \cite{li2021calibrating} proposed a paradigm, calibrating concepts and operations (CCO), which contained an executor to capture the underlying data characteristics for handling distribution imbalance and an operation weight predictor to highlight important operations and suppress redundant operations. 

	Requiring large scale, high quality, and labeled datasets are the hurdles of deep learning applications. To train an end-to-end neural-to-symbolic model, Agarwal et al. \cite{agarwal2021end} constructed a neural-symbolic-neural architecture. The model composed of three st.pdf providing a new fashioned way for image construction. Manigrasso et al. \cite{manigrasso2021faster} utilized logic tensor networks (LTNs) and a neural-symbolic technique to impose logical constraints on the training of the convolutional layers, which made the framework no longer purely data-driven. 
	
\begin{figure}[h!]
		\centering
		\includegraphics[width=1\linewidth]{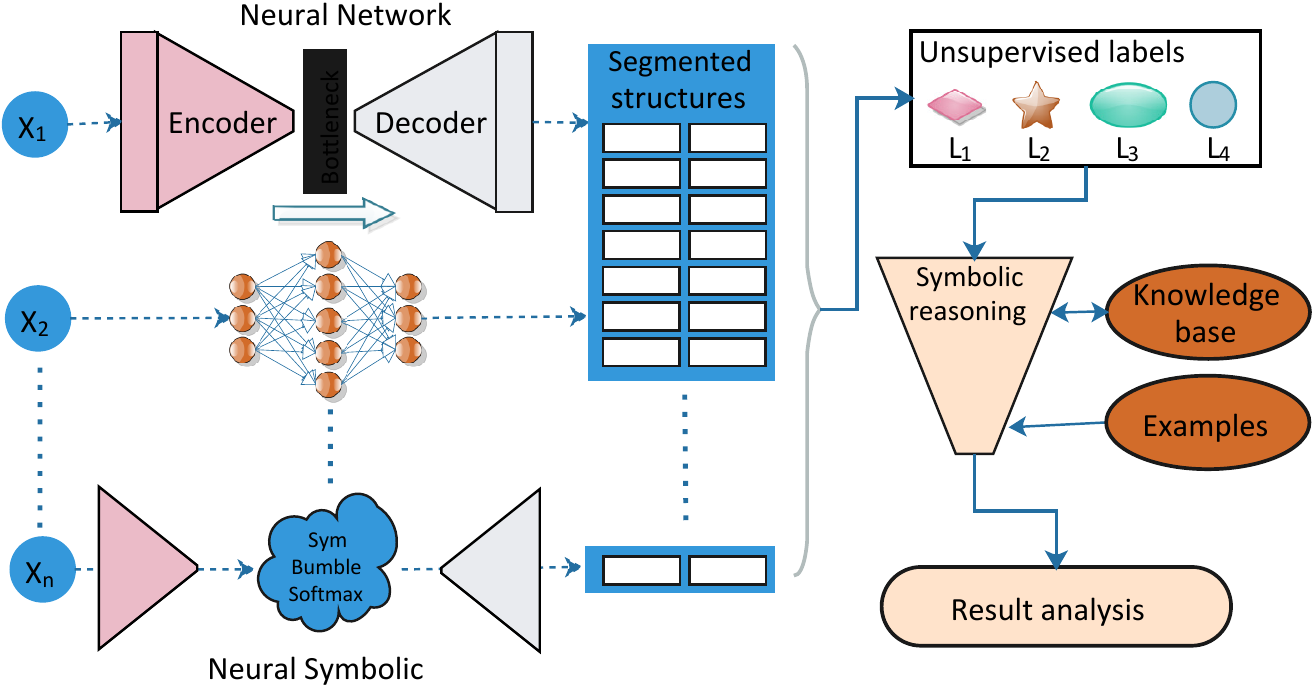}
		\caption{The combination of neural learning and symbolic learning. The neural network extracts features which passes for a symbolic unit for reasoning and inferencing. The symbolic reasoning feeds to an intermediate tensor composed of encoder-decoder structure for result analysis.}
		\label{fig:combining-NL-SL}
	\end{figure}

	The practical framework of neural symbols is summarized as shown in Figure~\ref{fig:combining-NL-SL}. The symbolic learning module embedded in the neural network can help .loklinterpretability and expalinability. Placing neural network structure before symbolic learning can make the system more robust in learning. 
	\subsection{The Possibility of Ocular Applications Using Neurosymbolic AI}
	Some efforts \cite{yi2018neural,marino2021krisp, mao2019neuro,liang2020lrta, vedantam2019probabilistic,bennetot2019towards} have been made to use NeSyL to assist deep learning representation to process VQA , bringing more transparency and explainability to the reasoning process.The neuro-symbolic visual question answering (NS-VQA)~\cite{yi2018neural} based model composed of three parts, scene parser, question parser, and program executor, to disentangle the reasoning of visual perception and textual comprehension. The core part designs symbolic programs to reason both visual and textual representations obtained through deep learning and question-answering in ocular clinical scenarios. The knowledge reasoning with implicit and symbolic rePresentations (KRISP) was proposed by ~\cite{marino2021krisp} to adopt two branches of reasoning, one is to integrate explicit symbolic knowledge into the knowledge graph for graph-based inference, and the other is to complete the supervised implicit knowledge. The look, read, think, answer (LRTA) \cite{liang2020lrta} treats the VQA task as a complete answer-generating question, proposing an end-to-end trainable modular framework that is not just a superficial guess to answer the question, but highly rationalizes throughout the process. 
	
	In addition, NeSyL plays a similar role in other reasoning tasks \cite{zhang2021abstract,feinman2020learning, liang2016neural, sen2021neuro, li2020closed}. The probabilistic abduction and execution (PrAE) Learner \cite{zhang2021abstract} uses a NeSyL concept to process Raven’s progressive matrices (RPM) like spatiotemporal inference tasks and to design probabilistic abductions to generate answers like humans. A generative neuro symbolic (GNS) model was proposed by \cite{feinman2020learning} to utilize compositional and causal knowledge of NeSyL to generate task-general representations of handwritten character concepts. Through the above research, it can be found that combined with interpretable NeSyL, deep representation learning can provide more transparent processes of reasoning while achieving satisfactory performance. Specifically, deep learning modules tend to be responsible for generating certain tasks followed by symbolic learning to complete the entire reasoning process and make it interpretable to human.
	
	The neuro-symbolic concept learning (NS-CL)~\cite{mao2019neuro} built on object-based scene representation and translation into an executable symbolic program can learn visual concepts and semantic information without explicit supervision. Neural networks build a knowledge base instead of manually extracting features, and the amount of data is only 10\% of that of neural networks. This outcome makes the neural notation accessible on images including all types of ocular images.
	
	\section{Innovative Ideas or Future Directions}
	NeSyL integrates contemporary neural network-based models with features of symbolic learning and reasoning. The research community is intrigued about how to make interpretable neural network models that transcend the black-box representation by incorporating properties like interpretability and explainability. This section offers novel concepts that may be applied to many fields in real-world situations. These concepts are both generic and specialized. This part is divided into two subsections: NeSyL-based generalized in clinical settings including medical imaging and the diagnosis of ocular diseases.
	
	\subsection{Domain Specific NeSyL for Medical Imaging}
	Given the challenges in neural computing in terms of transparency, interpretability, explainability, safety and trust, perturbation attack, and general robustness, bring commonsense to machine learning which is unavoidable for a reliable neural learning. In biomedical science, health, therapeutic practices, and reasoning in machine learning are becoming increasingly relevant. 
	
	\subsubsection{NeSyL for Multi-class and Fine-grained Classification of Ocular Diseases} NeSyL applications in clinical practices is advantageous considering the interpretability and explainability. It can be broadened to various medical imaging including fundus images, which can be utilized to diagnose up to 50 disease where each disease may have multiple subtypes. Among these disease, retina image classification is critical considering the subtle differences between sub-categories and large intra-class differences ~\cite{xiao2015application}. The signal-to-noise ratio of fine-grained images is very small where the information containing sufficient discrimination often only exists in very small local areas. Effective utilization of local area information has become the key to the success of fine-grained image classification algorithms. Moreover, the medical imaging, especially in the field of ocular diseases, often demands the involvement of the expertise for an expensive task of objects labeling~\cite{xie2015hyper}.
	
	The fine-grained image classification extracts objects with clear object shapes, outlines, appearance, and requirements of large scale reliable dataset. However, for objects with unclear contours and no obvious components, such as extracted lesions with different grades and minor differences, neural network algorithms may not achieve satisfactory performance or interpretability. On the other hand, NeSyL not only extract features via Connectionist AI methods, but also manipulate features using symbolic AI methods. The introduction of attention mechanisms address more in depth processing of extracted features, but still lack symbolic reasoning toward interpretablity and explainability models. Therefore, NeSyL has shown better performance, while requiring a small amount of training data, and the capability of learning and reasoning in performing distinct tasks. Thus, NeSyL has good application prospects in medical images specifically in the field of ocular disease classification.
	
	\subsubsection{Knowledge as Symbolics Regarding Clinical Practices} Knowledge-driven platforms for health-care decision-making are useful tools for giving actionable and intelligible information to patients and clinicians. Building knowledge into learning algorithm improves data efficiency and learning rates. Hierarchical knowledge or learning group sequence of related low-level actions into subgoals, where tasks are abstracted into subgoals for efficient learning. A capsule network can be employed to capture hierarchical relations and feature selection. To bring facilitation and reasoning about a patient's health, it is inevitable to integrate and utilize the domain knowledge to enable personalized health care system~\cite{shirai2021applying}. Symbolic information representation into domain knowledge can be focused to areas such as patient personal knowledge. The Patient personalized knowledge representation as a graph (PKG) encapsulates relevant knowledge about a patient. For instance, regarding a diabetic retinopathy patient, the patients' health, food, diet, lifestyle, and history are needed to model the health care decision system. The PKG about a person health's can be captured from various sources ~\cite{montoya2018knowledge}. The PKG in case of patient health can capture knowledge about patient history including allergies, eating habits, and other related diseases.

\begin{figure}[h!]
		\centering
		\includegraphics[width=1\linewidth]{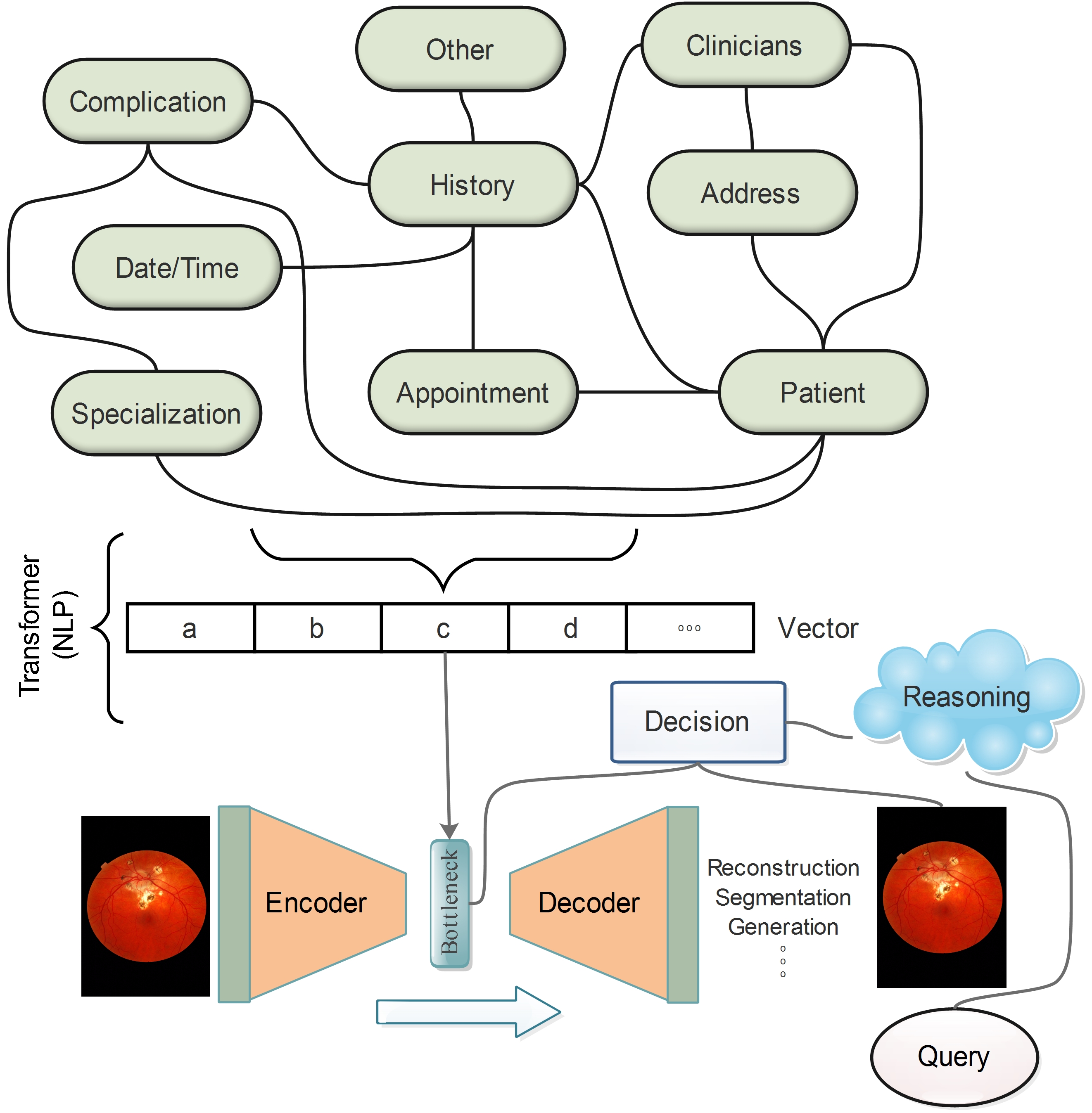}
		\caption{Personal knowledge as symbolic representation into an encoding-decoding structure of a neural network. The reconstructed, generated, and segmented image can be reasoned for decision making purpose.}
		\label{fig:PKG-NSL}
	\end{figure}

	Given the PKG as domain knowledge about a patient's health history through symbolics, a neural network could be trained to reason about the health informatics such as meal recommendation, allergic restriction, and carbohydrate prescription (Figure~\ref{fig:PKG-NSL}). The PKG knowledge can be translated into a graph or hypergraph prior to integration (symbolics) and inference with the neural network while training. The trained NeSyL network may not only predict patients diagnosing but also prescribe health care. It can be challenging to collect personal health information, linking or embedding such information to neural network, and most importantly associating against the clinical statistics of the same patient to map the full health history. Using machine learning in isolation may not reflect the efficient utilization of the domain knowledge. Thus, NeSyL can better inference, reason, and provide a health decision system akin to human clinicians.  
	
	\subsubsection{NeSyL Parallelism with Synthetic Biology} Synthetical biology transforms the technique of functional biomaterial production, leading to the development of new classes of inexpensive, and rapidly deployable diagnostics. \textit{AI} has been incorporated both in NeSyL and Synthetic biology (synbio). In genetics, AI has been employed to annotate protein function, heterologous genes optimization, estimation of plasmid expression, genotype-phenotype understanding and association. Furthermore, synthetic biology aims to control complex biological system by leveraging engineered principle and rules. The engineered rules standardize the genetic parts to achieve a projected outcome. Similarly, NeSyL utilizes, presents, integrates, and enables inferencing of symbolic knowledge in terms of domain knowledge with neural architecture. A veriety of AI approaches have been employed to Synbio such as multi-agent system, expert system, heuristic search, and constrained based reasoning~\cite{bilitchenko2011eugene,pedersen2009towards,yaman2012automated}. An attempt has been made to incorporate biological or domain knowledge into graph neural network ~\cite{ma2019incorporating}.

\begin{figure}[h!]
		\centering
		\includegraphics[width=1\linewidth]{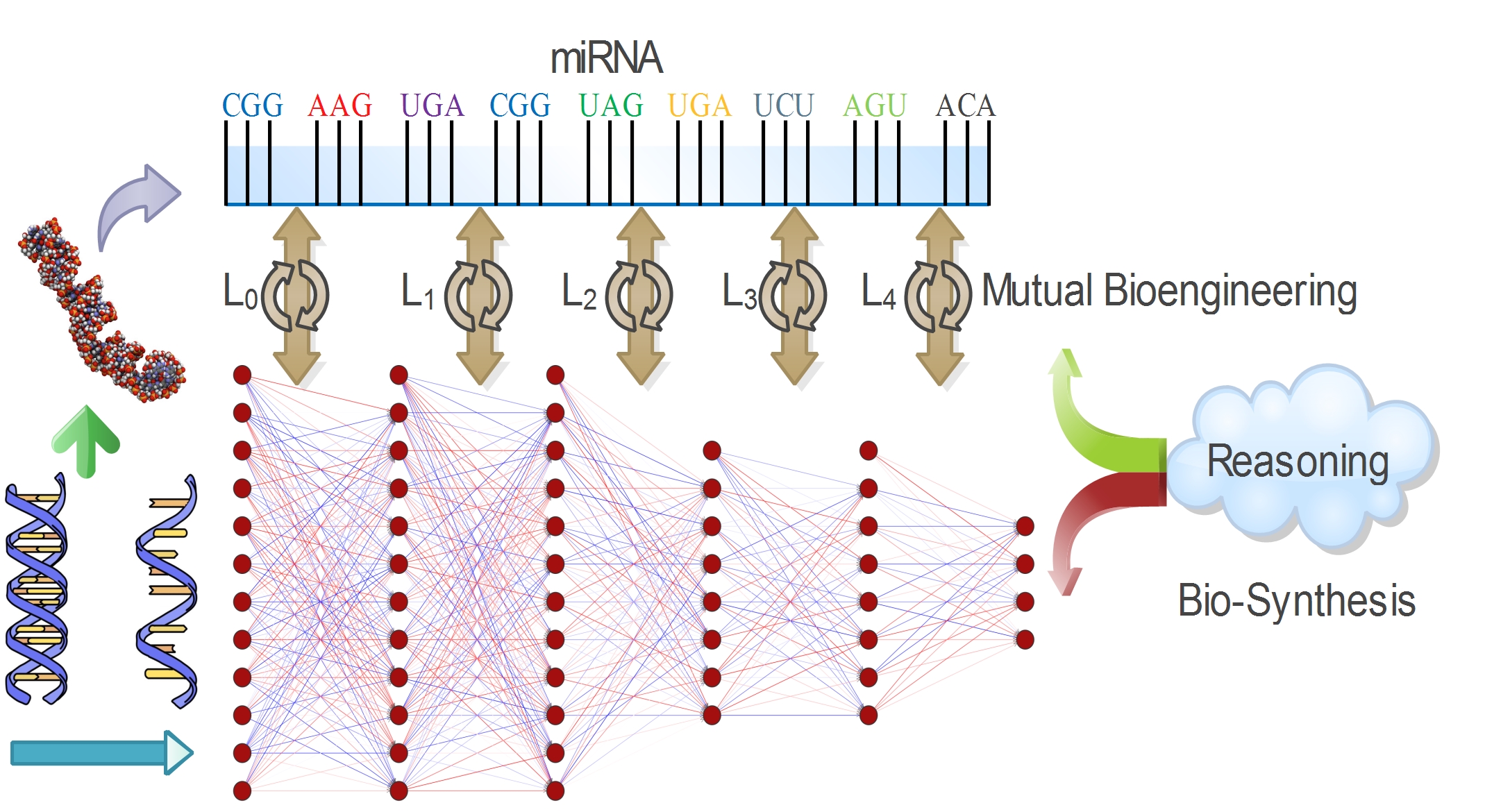}
		\caption{Mutual learning of bioengineering and artificial neural network. Both miRNA synthesis and neural network finetuning carry in parallel while learning at distinct levels as necessary.}
		\label{fig:Synth-Biology}
	\end{figure}

	Following the similarities of Synbio and NeSyL, a joint model can be built to reason at the genetic information level by utilizing biological knowledge and neural computing. This idea will not only engineer the genetic information but also has the potential to apply inferencing such as protein-protein interaction, DNA and RNA sequencing, identifying proteins functional sites, and the construction of new biological designs. Furthermore, there are a number of challenges that could be faced while modeling by incorporating symbolics into neural computing, for instance, dataset (lack of dataset to merge AI into synthetic biology, omics dataset), no robust algorithms (to capture system dynamics in high dimensional space), metrics (to account underlying biological system), and so on. NeSyL is the concept of imitating the human inferencing system while synbio deals with modifying genetics, thus their mutual collaborations would be an interesting future research direction. AI can help bioengineering, while in reverse, bioengineering principles and findings can help AI. Thus, by deploying NeSyL in synthetic biology, a gene can be modified at a gnomic level and inference for our desire outcome (cell self-repair and self-replication). A generalized model has been depicted in Figure~\ref{fig:Synth-Biology}, in which the miRNA synthesis and neural network (NN) perform in parallel as siamese networks. The miRNA editing utilizes biological knowledge which may compose a sequence of stages. Similarly, a neural network that is inspired by biological learning can tune network parameters for similar fed domain knowledge. Both the parallel systems can be used for reasoning and inferencing toward an optimized NeSyL approach. For instance, NN and Micro-RNA synthesis can mutually benefit from each other simultaneously and finally contribute to biosynthesis in genetic editing,  reasoning, and inferencing neural networks by forming a NeSyL platform.   
	\subsubsection{Diseases Diagnosis, Prognosis, and Treatment}
\begin{figure*}[h!]
		\centering
		\includegraphics[width=1\linewidth]{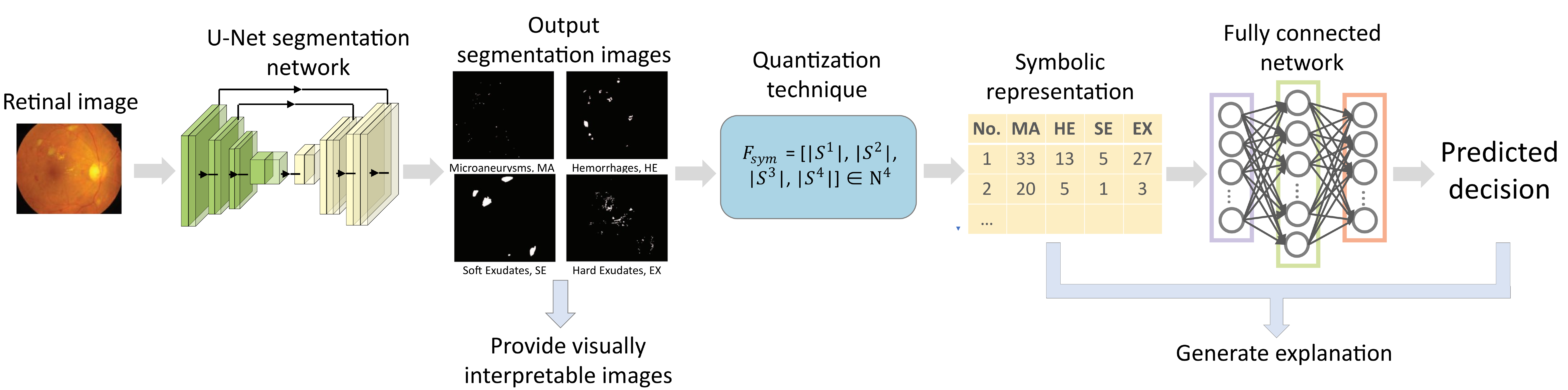}
		\caption{The ExplainDR utilizes a feature vector as a symbolic representation and achieves explainable and interpretable DR classification. The end-to-end architecture receives retinal images into a U-Net based segmentation network to generate four ouput images for each eye's condition. Then a quantization technique is designed to extract symbolic representation followed by a fully connected network to predict decision.}
		\label{fig:diagnosis_prognosis_treatment_DR}
	\end{figure*}

	 Explainable machine learning approach (ExplainDR) has been used to diagnose ocular disease (i.e., Diabetic Retinopathty)~\cite{jang2022explainable}. A generalized proposed architecture for diagnosing ocular diseases is shown in Figure~\ref{fig:diagnosis_prognosis_treatment_DR}. The study used static vector for symbolic information given 4 eye conditions (severity level) and statistically decides segmented lesion sizes (small, medium, and large). The accuracy increased from 47\% to 64\% given 4 to 12 dimensions of symbolic features. 
	
    The increases have been observed in accuracy corresponding to the growing number of symbolic features. Therefore, additional feature dimensions inclusion may make the framework significant; nevertheless, this may complicate the symbolic reasoning. In medical image diagnosis, prognosis, and treatment, the domain (expert) knowledge has much importance to be infused with the neural network at the different possible levels of integrations. The domain knowledge can be incorporated into the latent space and generating segmented regions simultaneously with symbolic information with reasoning capacity~\cite{neal2012bayesian, krupka2007incorporating}.
	
	In medical imaging practices, the acquisition of detailed information is indispensable for clinics in order to diagnose and treat diseases. Biomarkers or lesions could require detection, segmentation, classification, and spatial information extraction to correctly retrieve the information of interest. This may demand the employment of multi-tasking. The multi-tasking in this case would need to utilize the domain knowledge and symbolic reasoning to predict accurate results. Similarly, the segmented lesions may have variations in contrast and intensity that represent symptoms and their correlation is necessary for prognosis and treatment. The embedding of domain knowledge in terms of symbols needs special attention in neural computing, for instance, glaucoma estimation not only lies in cup-to-disk-ratio (CDR) but factors, such as age, race, and family history can also be utilized to carry out symbolic reasoning.    
	
	\subsubsection{Tele-Medicine}
	Telemedicine, especially in the field of ophthalmology, can improve the sensitivity of eye disease screening in communities or primary care settings and provide care to patients in resource-depleted areas~\cite{li2021digital}. Over the past few decades, there have been numerous teleophthalmology pilot projects around the world, many advanced vision algorithms can be introduced into this area given ophthalmology data as images. For instance, supervised and unsupervised ML has achieved new breakthroughs in automated glaucoma screening, proving the feasibility of telemedicine detection and management.

	The safety of remote triage in emergency ophthalmology remains to be explained and improved. An early study showed that 1\% of 500 patients undergoing teletriage in the emergency department experienced delays in treatment due to misdiagnosis~\cite{li2021digital}. A more accurate description on potential harms should be made before teletriage is widely available and establish mechanisms to mitigate the disadvantages of remote censorship. The advantage of NeSyL is not only to maximize the efficiency of the network but also to use symbolic reasoning to improve the interpretability of pathology and make the model prediction more accurate, which has far-reaching significance for ophthalmic remote triage research.
	
\begin{figure}[h!]
		\centering
		{\includegraphics[width=0.5\textwidth]{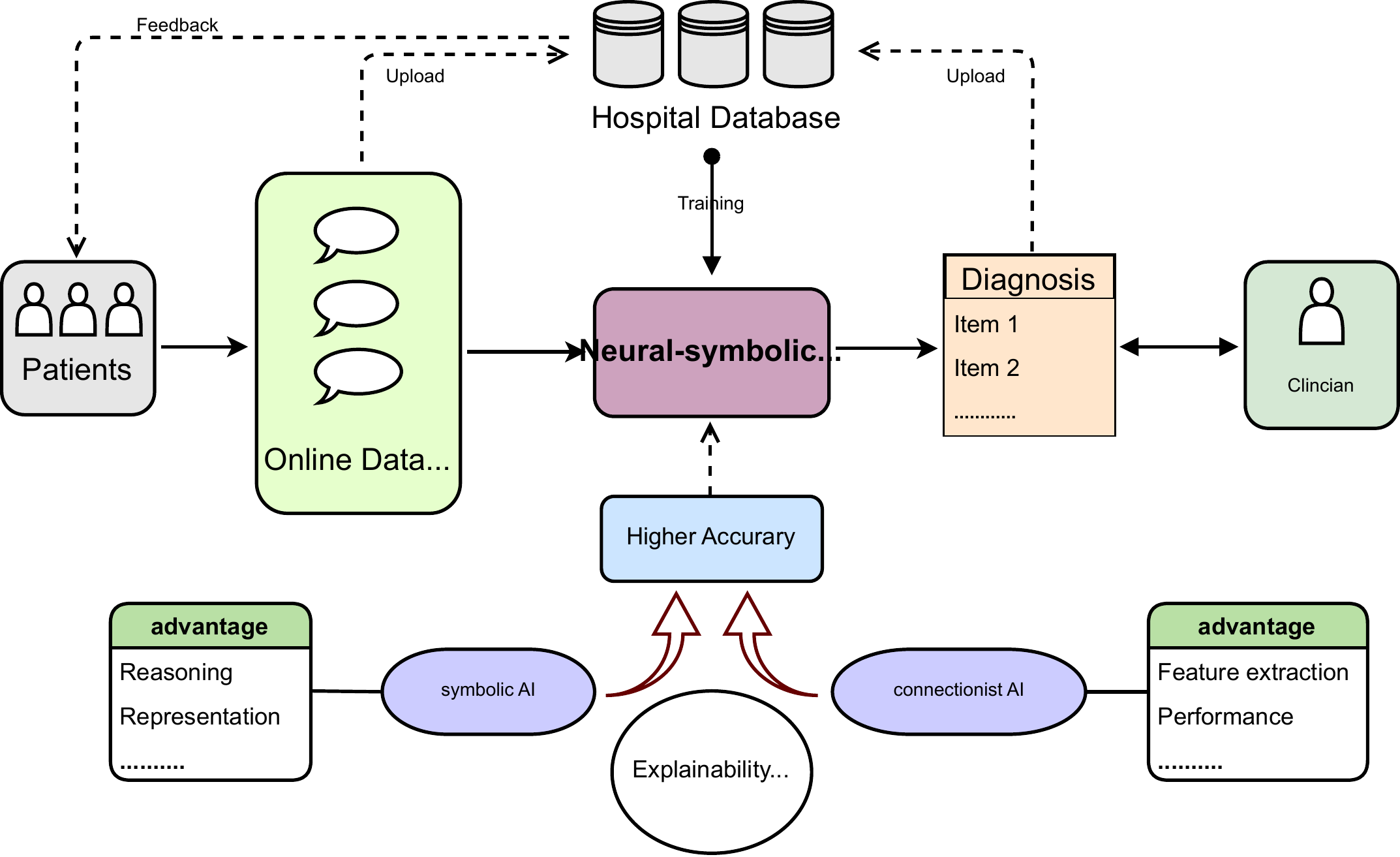}
		}
		\caption{Smart healthcare telemedicine service.}
		\label{fig:Smart-Health-Telemedicine}
	\end{figure}

	Figure~\ref{fig:Smart-Health-Telemedicine} shows a tele-ophthalmic medical diagnosis system that is based on NeSyL, and the trained model can achieve more accurate diagnosis with the advantages of neural network and symbolism (NeSyL). The pathological data of patients is obtained online, and the trained model makes a detailed diagnosis conclusion, then the clinician supervises its accuracy, which greatly reduces the miss-diagnosis rate.
	
	\subsection{Innovations in Broad Perspectives}
	\subsubsection{Addressing ML and DL Challenges in NeSyL}This section covers and leads experiments using NeSyL to address issues in recent machine and deep learning approaches.
		
	\textbf{Addressing black-box challenge through NeSyL:} The ability to comprehend a machine learning model's prediction builds user trust and aids understanding of the underlying processes. In terms of interpretability and reasoning, one of the major challenges faced by ML and DL-based modeling is the lack of a white-box idea. Model interpretability is important in a variety of application domains, including natural sciences, where understanding and prediction are important for decision-making \cite{alaa2019attentive, schmidt2009distilling,wang2019symbolic}. In this regard, some studies have been carried out to convert the black-box challenging aspect to a white-box characteristic in the form of visible mathematical equations that human subjects can readily understand and evaluate~\cite{alaa2019demystifying}.
	
	 \textbf{Optimization:} Varieties of approaches have been proposed to optimize a network parameters and hyper parameters~\cite{chen2019learning}. In discrete spaces where comprehensive enumeration is impossible, combinatorial optimization of symbolic learning seeks to identify optimal configurations. Nonetheless, despite a worst case exponential time complexity, there is a wide range of accurate combinatorial optimization algorithms that are guaranteed to discover an optimal solution~\cite{conforti2014integer}. Branch-and-bound, which is a classic precise approach for solving mixed-integer linear systems, was reformulated as a Markov decision process~\cite{gasse2019exact}. In this context, the study proposed and tested a novel approach for tackling the branching problem that involves expressing the state of the branch-and-bound process as a bipartite graph, which eliminates the need for feature engineering by leveraging the variable constraint structure of MLP problems naturally, and allows for the encoding of branching policies as a graph. A Knowledge distillation model might be employed to reduce the NeSyL model’s complexity so that it can be efficiently fine-tuned.
	 
	 The study in~\cite{oh2019combinatorial} proposed a bayesian optimization method for combinatorial search spaces. Most of the baysian optimization methods are focusing on continuous rather than combinatorial search spaces~\cite{movckus1975bayesian}. The innovated frameworks transform discrete constraints into smooth functions that are differentiable and optimized using gradient descent. To efficiently tackle the exponentially increasing complexity of combinatorial search space, the study ~\cite{oh2019combinatorial} represented the search space as the combinatorial graph by combining sub-graphs given to all combinatorial variables using the graph cartesian product. Moreover, the combinatorial graph reflects a natural metric on categorical choices (Hamming distance) when all combinatorial variables are categorical. Graph neural networks can be used to solve combinatorial optimization problems with a relaxation strategy to the hamiltonian problem to generate a differentiable loss function with which they train the graph neural network.
	 
	 \textbf{Conditioning on Neural Network:} Conditioning in the neural network aims to derive the process or signal among many forms toward the intended direction. The conditioning on neural networks have been carried out in varieties of studies, such as text-to-speech~\cite{oord2016wavenet}, style generation \cite{karras2019style}, attention mechanism \cite{xu1711fine}, literature generation, and text-to-image synthesis \cite{liu2021divco}. Humans innate ability for reasoning in terms of conditioning, such as textual description based visualization, is attempted in neural networking \cite{liu2021divco}. A hypergraph based conditioning can be applied at different levels, such as sentence-level and word-level conditioning for text-to-image synthesis, to bring flexibility in generation. The embedding of hypergraph as conditioning facilitates both the continuous and discrete image generation beyond the training dimensions. Furthermore, meta internal learning is deployed for image synthesis and generation (GAN) as condition on the network statistics, such as one-hot vector, vector signature, conditional signal, style as condition, and conditioning hypernetwork (rather than a primary network)~\cite{bensadoun2021meta}. Similarly, hypernetwork based variational autoencoder (VAE) has been conditionalized for self driving vehicle to predict the stochastic behavior of the multi-dimensional agent in novel environment featured with context driven and generalized scenarios~\cite{oh2022cvae}.

	 It is a genuine source to reason the neural network learning by imposing distinct conditions toward the intended goal. For example, the text-to-image synthesis which is crucial to learn from unstructured descriptions and encoding different statistical properties of language inputs. The future work can be aimed to apply controllable conditional (both discrete and continuous) based generation given hyper network parameters \cite{liu2021divco}. Furthermore, improvement can be brought by incorporating adversarial hypernetwork and meta learning such as few-shot learning \cite{bensadoun2021meta}. Implicit learning can be one of the future directions, in which the overall solution is based on sub-solutions, for reasoning in machine learning models~\cite{gu2020implicit}
	 
	 \begin{figure*}[h!]
	 	\centering
	 	\includegraphics[width=1\linewidth]{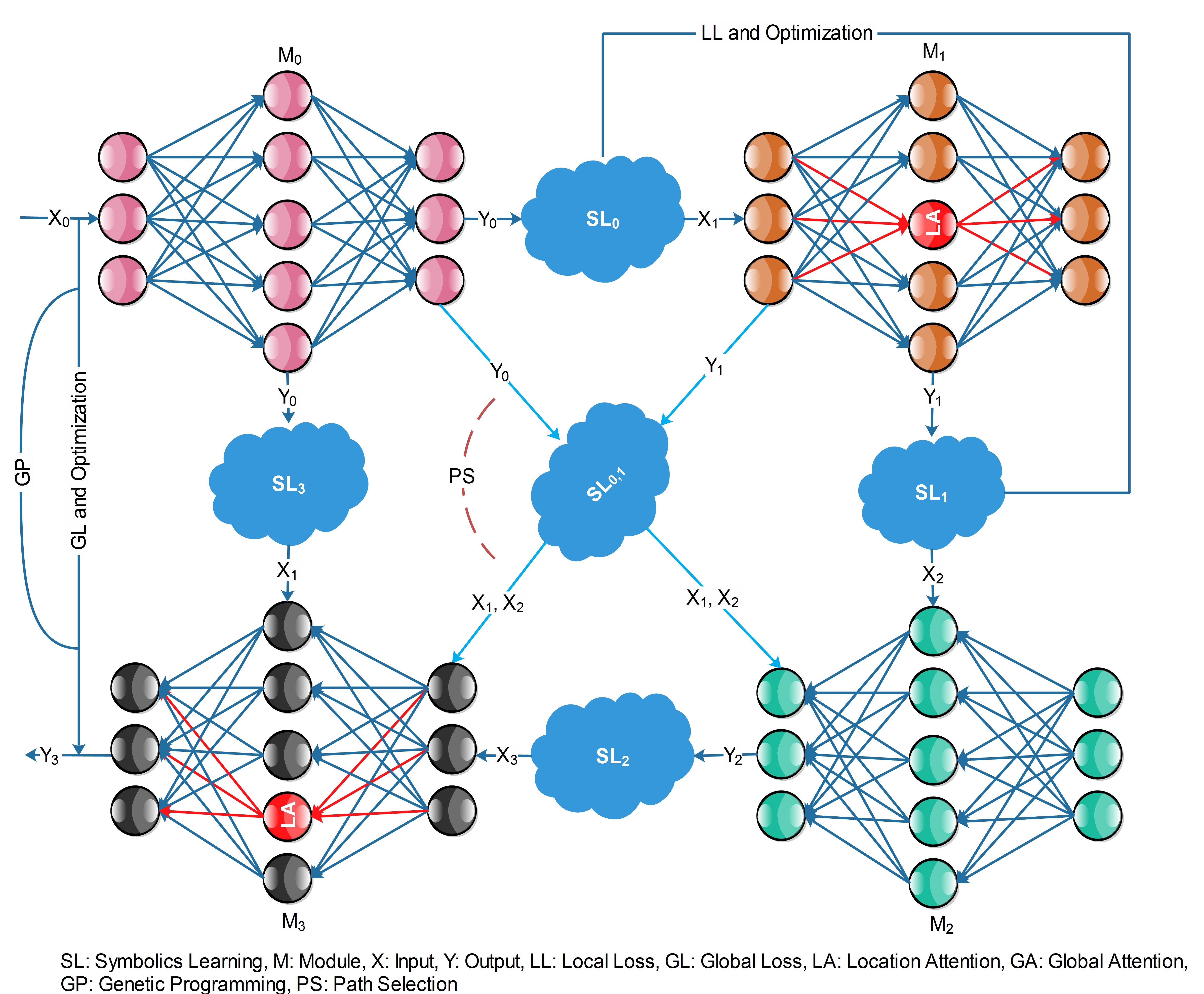}
	 	\caption{The figure depicts a generalized model toward interpretabiliyt, explainability, optimization, and condition on neural computing. The model presents four modules (M$_0$, M$_1$, M$_2$, M$_3$), symbolic learning submodules (SL$_0$, SL$_1$, SL$_2$, SL$_3$, SL$_{0,1}$, SL$_{0,1}$), local (LL) and global losses (GL), local and global optimizations, and pathway selection (PS) using genetic programming (GP).}
	 	\label{fig:Black-box-Optimization-Conditioning}
	 \end{figure*}
	 \textbf{Discussion on Black-box to White-box Conversion, Optimization, and Conditioning:}
	 Regarding interpretability and explainability toward white-box learning, we proposed a generalized approach as depicted in Figure~\ref{fig:Black-box-Optimization-Conditioning}. The proposed model is based on NeSyL which address the hindrance of machine learning in terms of black-box representation. Overall, the model's structure has been divided into two layers (which can be extended as per requirement) including deeper layer (neural computation) and upper layer (symbolic learning). Each submodule in the deeper level receives input (can be symbolics) and learns parameters in neural network architecture and then produces output to a symbolic learning (SL) unit. The proposed approach deals the black-box to white-box interpretation by incorporating optimization and conditioning (BWOC). The key idea is to divide the whole end-to-end ML platform into submodules infused with symbolics at different levels. The architecture of BWOC carries out NeSyL at low level to overcome the abstraction and enable understanding neural network to achieve white-box architecture. The BWOC still has the abstraction up to some extent to address the phenomena like adversarial attack. Thus, BOWC provides the opportunity to include and exclude modules given their performances while learning in a particular domain.

	 The BOWC optimizes the parameters both at local and global levels (Figure~\ref{fig:Black-box-Optimization-Conditioning}). The local loss (LL) enables the optimization at particular module when realized to be higher than fellow modules. Similarly, the model also cumulatively computes the loss function and optimizes in an end-to-end manner. The global optimization can be carried out by selecting a module or number of modules that have high a contribution to performance. The selection of the optimized module can be performed using genetic programming (GP). Meijer G-function can be used for selection purposes given symbolic expressions such as polynomial, arithmetic,
	 algebraic, closed-form, and analytic expressions, as well as some special function including hypergeometric and bessel functions. One of the key aspects of the model is the path selection (PS), in which a global optimized path to be selected while training the model. 
	 
	 The third key aspect of the proposed model is the customization which can be in terms of embedding information and applying attention mechanism (AM). For the natural language processing, the embedding can be carried out using word, sentence, and structural levels. The AM can be employed both at global and local levels. For global level attention, particular module with low performance can be reparameterized given the learned knowledge by the optimized module. The PS can facilitate global attention mechanism while choosing a robust path. The local attention mechanism (LA) penalizes and prioritizes particular neuron(s) in a submodule level (see red colored neurons in M$_1$ and M$_3$ of Figure~\ref{fig:Black-box-Optimization-Conditioning}). In this way, the BOWC module following NeSyL concept exploits black-box learning, enable optimization at different levels, and infusion of different conditions toward a robust model. 
	 
	\subsubsection{Symbolic Reasoning}
	NeSyL is mainly focused on the typical machine learning problems with employing reasoning at the symbolic levels. There is no systematic way for reasoning which need more attention to reach a maturity level. It is necessary to propose innovations in reasoning by achieving human-level reasoning. A number of reasoning methods have been proposed, such as object-centric~\cite{chen2021grounding} (reasoning about object and its attribute), logical reasoning~\cite{garcez2019neural} (combining expressive NN with symbolic reasoning), neuro-symbolic forward reasoning-NSFR~\cite{shindo2021neuro} (differentiable forward-chaining using first-order logic reasoning), common sense reasoning~\cite{arabshahi2020conversational}, and visual question answering~\cite{amizadeh2020neuro}. 
	
	The object centric suffers from the complexity of performing low-level visual perception while reasoning on upper or higher level, which has been addressed by employing logical reasoning (symbolic representation) with deep learning~\cite{garcez2019neural}. A dynamic concept learner approach has been deployed to not only capture features about objects but also to track trajectories of each object movement~\cite{chen2021grounding}. Similarly, adaptive neurosymbolic network adapts to its environment by building symbols based on perceptual sequences. 
		
   For building intelligent systems, intuitive learning (intuitive physics) has been utilized for machine cognition to reason with the physical world having incomplete information~\cite{duan2022survey}. The reasoning at physical level can be divided into \textit{prediction}, \textit{inferences}, and \textit{causal}. The physical level reasoning use DL modeling to outcome physical trajectories, dynamics, and objects properties. In inferencing, the observed (e.g., size, color, and shape), latent properties (e.g., mass, friction, velocity, displacement), and inter related in the dynamics such as collisions, bouncing balls, and objects moving to utilize to reason and inference. The generation of events, movements, dynamics, and unseen future frames can be forecasted given the initial dynamics.

	The machine cognition and reasoning with uncertain physical and incomplete physical world and dynamics, a bayesian symbolic framework (BSP) has been proposed~\cite{xu2021bayesian}. The BSP approach combines the sample efficiency of symbolic methods with the accuracy and generalization of data-driven approaches to incorporate the learning and reasoning (like a human) from incomplete information. In this regard, parse-tree-guided reasoning network achieved global reasoning on a dependency tree parsed from the question capable of building an interpretable VQA system that gradually derives image cues following question-driven parse-tree reasoning. 
	
	The existing methods in intuitive reasoning do not generalize from one scenario to another, for instance, the generalization of two objects collision to multiple objects' collision simultaneously. Similarly, in intuitive physics, the inferencing of latent properties (mass, density, friction, energy, velocity, displacement, etc.) is challenging to extract features from the observed properties (size, shape, and color) by human intervention. Thus, the mapping or derivation of latent properties from the physical world need to be modeled. The inferencing can be modeled by separating visual properties from dynamic properties in the physical world. Similarly, for inferencing, a bayesian symbolic framework has been proposed which combines inferencing with symbolic regression to address the uncertainties in physical dynamics~\cite{xu2021bayesian}.   
	
	For future studies of reasoning and inferencing specifically in intuitive physics and cognition, understanding dynamics (e.g., objects over time, collision, direction, bouncing, gravity, etc.,) and their proper representation (e.g., graph~\cite{battaglia2016interaction}) are inevitable~\cite{chen2021comphy}. To understand the physics of a scene, it is necessary to reveal the compositional hidden properties, such as mass and charge, and to utilize these factors to improve the symbolic reasoning. In the physical world, the information may be incomplete to model dynamics. Therefore, to understand an environment, all the newtonian forces, physics laws (learnable symbolic expressions using meaningful law), formulation, and entities forces on each other, together with data driven (DL) can be integrated and reasoned about incomplete or intuitive information. 
	
	Most of the current reasoning approaches have been applied on images and videos which can be extended to 3D scenarios with sophisticated algorithms. Similarly, neural network can be used for reasoning considering its strong backbone for reasoning such as dynamic computational graph~\cite{chen2021grounding}. The reasoning is naturally linked to language, which makes natural language processing (NLP) appropriate for NeSyL. However, whether NLP and NeSyL mutually benefit from each other is still an open question.  
		
\begin{figure*}[h!]
		\centering
		\includegraphics[width=1\linewidth]{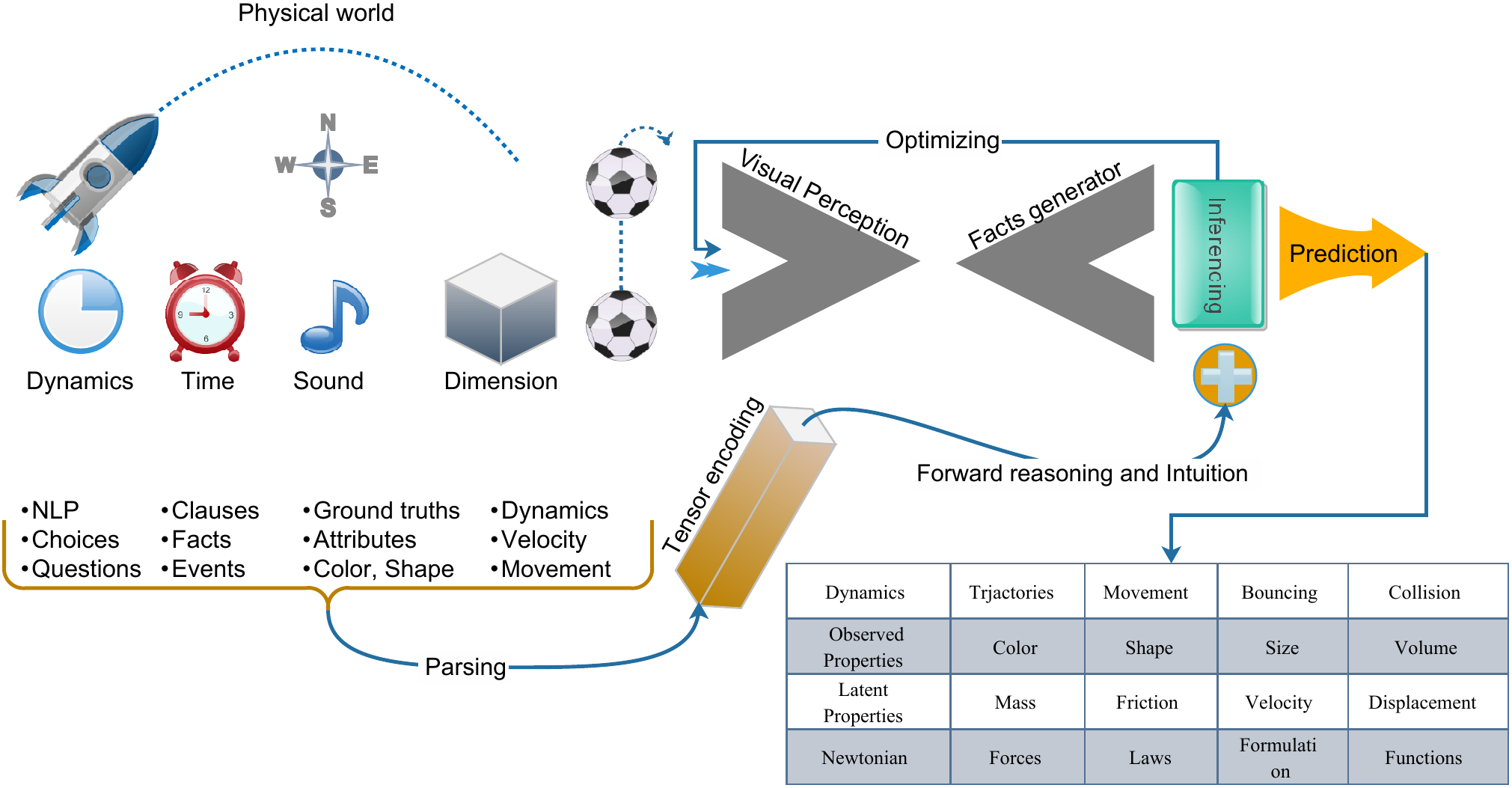}
		\caption{The proposed NeSyL model receives visual perception and observed information such as color, shape, size, and volume together with tensor encoded information to reason and inference with training model. The model utilizes the forward reasoning and intuition about observed scenes in the inferencing phase where the inferencing phase facilitates back the network optimization. Finally, the model predicts both observed and unobserved information as tabulated.}
		\label{fig:Symbolic-reasoning}
	\end{figure*}

	Therefore, we proposed a generalized NeSyL based model to intuitively reason and inference from the physical world via symbolic learning toward incomplete, hidden, latent, and intuitive information (Figure~\ref{fig:Symbolic-reasoning}). The proposed approach receives visual information from physical scenarios and perceives knowledge in a neural network in terms of weights learning. The neural part of the model generates facts given observed information acquired in the physical scenes. Typically, a physical world may contain uncleared and incomplete information for deducing the dynamics. Similarly, a scenario cannot be generalized to other scenarios. The proposed model receives additional information as symbolics in terms of natural language, clauses, ground truths (observed information), and dynamics passed through a parser into a tensor representation. Such representation facilitates the model prediction for intuitive information and also optimizes the weights. The model finally predicts both kind of observed and derived information given the factual visuals. Most importantly, the NeSyL solving approach to incomplete and intuitive information can be extended into medical practices and opthalmology with attaining life saving results.  
	
	\subsubsection{NeSyL for Interactive and Critical Systems} Despite of the wide range of applications and efficiency, DL models can still be fragile and sensitive to given data. In the adversarial attack, adding imperceptibility noise to an image can easily fool a DL based system, such as by classifying a gibbon as panda~\cite{papernot2017practical}, fooling adversarial network with different lightnings and orientations~\cite{kurakin2018adversarial}, dodging face detection with stealthy attacks~\cite{sharif2016accessorize}, hiding in the human eye as graffiti~\cite{eykholt2018robust}, and sticker and patch adaption based network tricking~\cite{brown2017adversarial}. Such vulnerabilities of generative models may get worse in a real-time and interactive system where a minor error can lead to huge consequences. For instance, a model for autonomous cars trained on the road side visuals may be mislead by observing perturbation on the perceived visuals. Some of the autonomous vehicle established in industries, including Tesla (HydraNet), Waymo (ChauffeurNet), and Ford/Volkswagen Group (Argo AI) are avoiding end-to-end system to better identify and fix problems while dealing with challenging environment. 
	
	Therefore, the futuristic autonomous vehicle can be equipped by combining typical machine learning with tracking and commonsense facility. A number of studies have been attempted to combine sensory information, objects recognition, and tracking with symbolic learning which has improved the object detection from 0.31 intersection-over-Union (IOU) to 0.65 in the simulated environment~\cite{yun2022neurosymbolic}. Similarly, the interactive systems must consider dynamics such as object projection and interpolation (next step based on the occluded information), identity maintenance (error detection and tracking), objects attribute assumption, commonsense, and counterfactuals (mimicking human cognition). For this purpose, a visual sensemaking framework has been proposed to integrate knowledge representation and computer vision for interactive system \cite{suchan2021commonsense}. 
	
	However, NeSyL could be made more robust and efficient via integration of multi-sensory information, multi-agent merging, contextual knowledge, and situational information in modular framework. One of the key areas that need to pay more attention is the medical practices including opthalmology to produce a lifesaving outcome. 
	
	\subsubsection{NeSyL Integration and Embedding} Considering the integration of symbolic learning into the neural network algorithms, varieties of integration methods have been employed following the essence of the problem~\cite{chaudhuri2021neurosymbolic}. However, there is no systematic, generalized, and standard approach to integrate symbolic (high level) and subsymbolic (neural network level) representations. A taxonomy of NeSyL is viewed from two perspectives: the coupling magnitude (strong, tight, and weak) and coupling types (pre/postprocessing, subprocessing, coprocessing, and metaprocessing). In this section, we elaborate some existing approaches and propose futuristic NeSyL architectures to infuse domain knowledge with neural network learning. The following sections illustrate related concepts.
	
	\textbf{Graph and Hypergraph based Integration:} The integration via graph and hypergraph provides an effective capturing relationship between symbolic and sub-symbolic (neural network level), independency on programming language, and most importantly, preserves the topological dependency of information \cite{latapie2021metamodel}. Graph representations are inherently compositional in nature and allow us to capture entities, attributes and relations in a scalable manner. The study \cite{saqur2020multimodal} proposed a multimodel graph neural network to induce a factor matrix for question answering session and to solve the problem of compositional generalization for visual reasoning. 

	Knowledge graphs, for example, frequently display numerous hierarchies at the same time~\cite{balazevic2019multi}. A spatial network, such as hyperbolic space, may be used to depict a data structure that is a continuous imitation of discrete trees, allowing it to be used to model hierarchical and graphical data. The advantages of hyperbolic-based embedding include fewer dimensions and promising results in downstream applications.

	\textbf{Layer-wise Integration and Attention:} In the study ~\cite{latapie2021neurosymbolic},  varieties of stressing (attention) mechanisms are employed between the symbolic and subsymbolic levels. Similarly, different layers of abstraction have been utilized to integrate high and low level information in terms of symbolic and subsymbolic \cite{latapie2021metamodel}. The attention mechanism among these layers (L$_0$, L$_1$, ..., L$_n$) should be cognitively synergized among all the layers.  Thus, sophisticated methods are needed to deal with attentions originating from symbolic and subsymbolic systems and should be assured their coordination, synergization, harmonization, and coherence in interactions. Attention mechanism has high significance of modeling NeSyL, such as particular neurons activation in the neural system by perceiving certain symbols \cite{velik2010neurosymbolic}. 
	
	The study~\cite{latapie2021metamodel} proposed a deep fusion based reasoning engine (DFRE) to represent the knowledge into four layers, where L$_0$ is close to the raw sensor data obtained from physical system. This study attempts to represent, integrate, and reason knowledge mimicking human approach. The layer-wise integration has also been divided into physical, inferencing, and causal level \cite{duan2022survey}.   

	\textbf{Position-wise Integration of Domain Knowledge} In~\cite{dash2022review}, the domain knowledge inclusion has been studied at different positions such as input level, in the loss function, and as bias and weights parameters. The objective is to equip ethics, fairness, and explainability toward a responsible and stable neural network given the domain knowledge. If the input contains raw features with details retrieved from the domain, then the input would be transformed (propositionalisation) and fed to a DNN model. The domain knowledge can also be infused into the loss function as penalty term reflecting domain knowledge based guidance and directives. These terms may include the regularization-term, embeddings (low-dimensional learned continuous vector representations of discrete variables), and semantic domain constraint (specific restrictions on the prediction). Similarly, the domain knowledge can be incorporated into a deep network by introducing constraints on the model parameters (weights) or by making a design choice of its structure. Such incorporation can be explicitly performed in Bayesian formulation~\cite{buntine1991bayesian} as prior knowledge in the latent representation~\cite{neal2012bayesian, krupka2007incorporating}. The domain knowledge or symbolic information can be transferred to DL model using transfer-learning as prior distribution over the model parameters~\cite{wang2018deep}. Furthermore, domain-knowledge encoded as a set of propositional rules and domain constraints must be applied on the structure of the neural networks to achieve the intended goal~\cite{towell1990refinement,xie2019embedding,xu2018semantic}.  Similarly, domain knowledge can also be represented by metapaths as the integration of biological objects in a latent space.
	
	\textbf{Translational Integration:} There is no systematic way to translate logical constraints as symbolic representation while translating to neural network. The logical symbolic representations are not differentiable where the weights learning and propagation gradients similar to DL models training may not be carried out~\cite{evans2018learning}. To introduce the cycling dependencies in the symbolic representation, special attention, such as hypergraph, may need to translate into neural network. However, practically it is challenging to employ numerical constraints. The domain knowledge incorporation as loss term is not straightforward and not suitable for critical scenarios as well as special mathematical tool is needed to carry out the optimization. Furthermore, DL models confined to symbolic learning may not be scalable to large datasets.
	\begin{figure*}[h!]
		\centering
		\includegraphics[width=1\linewidth]{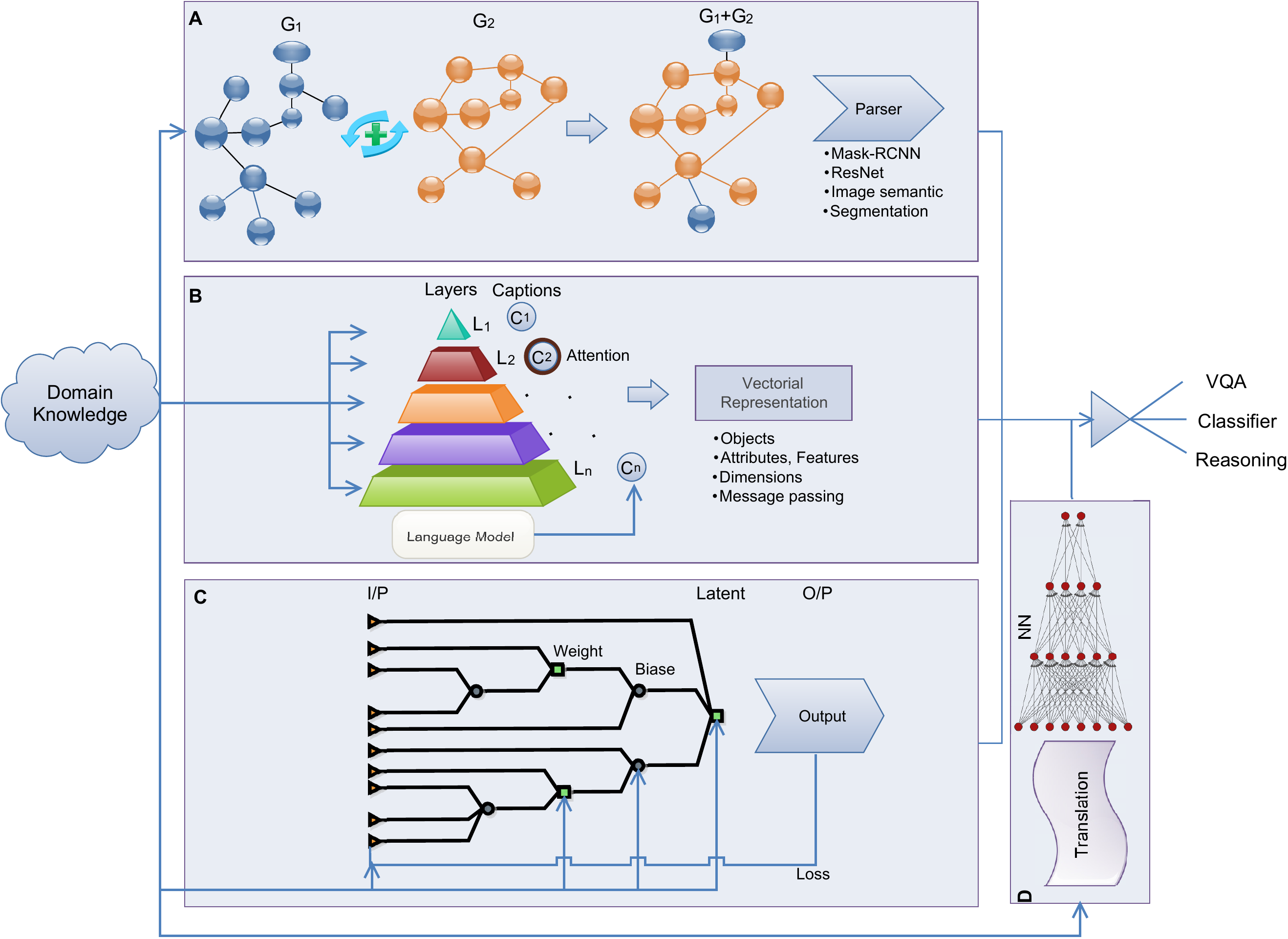}
		\caption{The domain knowledge integration in distinct ways has been depicted. (A) In the first type, the domain knowledge translate into graph and hyper-graph presentation to reflect complex relationship followed by a Parser. (B) The domain knowledge passes to a pyramid of different abstraction layers together with the corresponding caption and attention are needed (i.e., C$_2$). (C) Integration of domain knowledge at input level, latent level, or as weights, bias, and objective function. (D) Translation of symbolic knowledge into neural network.}
		\label{fig:NeSyL-Integration}
	\end{figure*}
	
	\textbf{Discussion on NeSyL Integration and Embedding:} The translation and infusion of domain knowledge must be confined with logical constraints prior to translating into the neural network. There is no standard framework for logical constraints that DL can consume which needs special attention prior the integration modeling. The symbolic logics are not differentiable like neural network which lead to the emphasis on making a unified modeling of domain knowledge with neural network. Moreover, there is no straightforward way to construct loss term regarding critical and safety circumstances given domain knowledge. 
	
	The graph-based approaches can enable smoother scaling and better manage complex fine-grained multimodal reasoning, and they can be used in applications such as robotic control and autonomous navigation. The inclusion of heterogeneous graph based networks may overcome the challenges of fairness in the treatment of complex relationships and interpretability~\cite{gu2020implicit}. 
	
	This study introduces four methods to integrate domain knowledge as symbolics into neural network (Figure~\ref{fig:NeSyL-Integration}). In the first method (Figure~\ref{fig:NeSyL-Integration}(A)), a complex environment of symbols can be presented in graph and hypergraph structure to transfer into the neural network architecture. Different levels of graphs can be generated given the input information where a cumulative graph sum up at the end prior to parsing. For instance, a physical world can be divided into different level of understanding, such as level-1 (G$_1$) can be included objects shape, appearance, design, and color, level-2 may include dynamics (G$_2$), where level-3 may include caption, source, and text. The dynamic world can further be divided into  rolling, floating, trajectory, and bouncing. Graph and hypergraph can better represent such complex structures. A parser can be used to extract features and to proceed for reasoning and inferencing purposes.   
	
	The domain knowledge can also be integrated into a layer-wise neural network architecture (Figure~\ref{fig:NeSyL-Integration}(B)). Each layer separately receives labeled information in terms of caption where specific layer has been pointed out via attention mechanism. The caption information has been received from language modeling. The layer based knowledge is then translated into vectorial representation comprising objects, attributes, dimensions, and message passing mechanism among layers. Moreover, the domain knowledge can be integrated at different position such as at input level, as biase, weights, and loss term (Figure~\ref{fig:NeSyL-Integration}). The domain knowledge can also be embedded by carrying out translating symbolics into neural network architecture (Figure~\ref{fig:NeSyL-Integration}). This paved the road to a systematic symbolic translation, addressing differentiability, and logical constraints representation into the neural network world.   
	 
	\subsubsection{Specialized DL Networks of NeSyL}
	Typically, the NeSyL uses symbolic learning (reasoning level) on top of neural learning (closely to physical world). The neural leaning can be reconstructive, generative, reinforcement learning model, and so on. However, little attention has been given to specialized DL algorithms for NeSyL. In the sudy \cite{zhan2021unsupervised}, unsupervised model based on VAE framework has been proposed, which encodes domain specific symbolic information or domain symbolic language coupled with neural network encoder to make the latent space interpretable and factorized. 
	
	The key idea is to use neural network and deep learning to generate the knowledge, and then the reasoning can be done in symbolic AI framework. And the knowledge dataset is very important for the symbolic learning. So, it is a big challenge on how to make the AI learns the knowledge as much as possible with as few as possible necessary manual input of knowledge data.	The encoding of domain knowledge into latent space by making cluster could be more challenging for multiple clusters and their simultaneous interpretation. The coupling of symbolic domain knowledge with standard encoder can enable human to interpret latent space corresponding to different applications such healthcare, sports analytics, and autonomous driving. However, such coupling should be avoided from any human attack as human are involved in data curation, domain knowledge designing, and interpretation.  
	
	Special attention is needed while designing DL model to embrace the novel features of symbolic learning in terms of integration, translation, reasoning, and application. So, the future DL models could be equipped with features of both incorporating domain knowledge and reasoning (temporal, structural, relational). Besides from the application perspectives, the pure image computational tasks, such as, image generation (synthesis), reconstruction, restoration, inpainting, registration, enhancement, and super-resolution, are also need to be modeled via NeSyL. Similarly, revisiting image processing scenarios including lacking annotation (zero-shot learning), absence of perfect and high quality as ground truth, and so on, with innovative approaches via NeSyL. 
	 
\begin{figure}[h!]
		\centering
		{\includegraphics[width=0.5\textwidth]{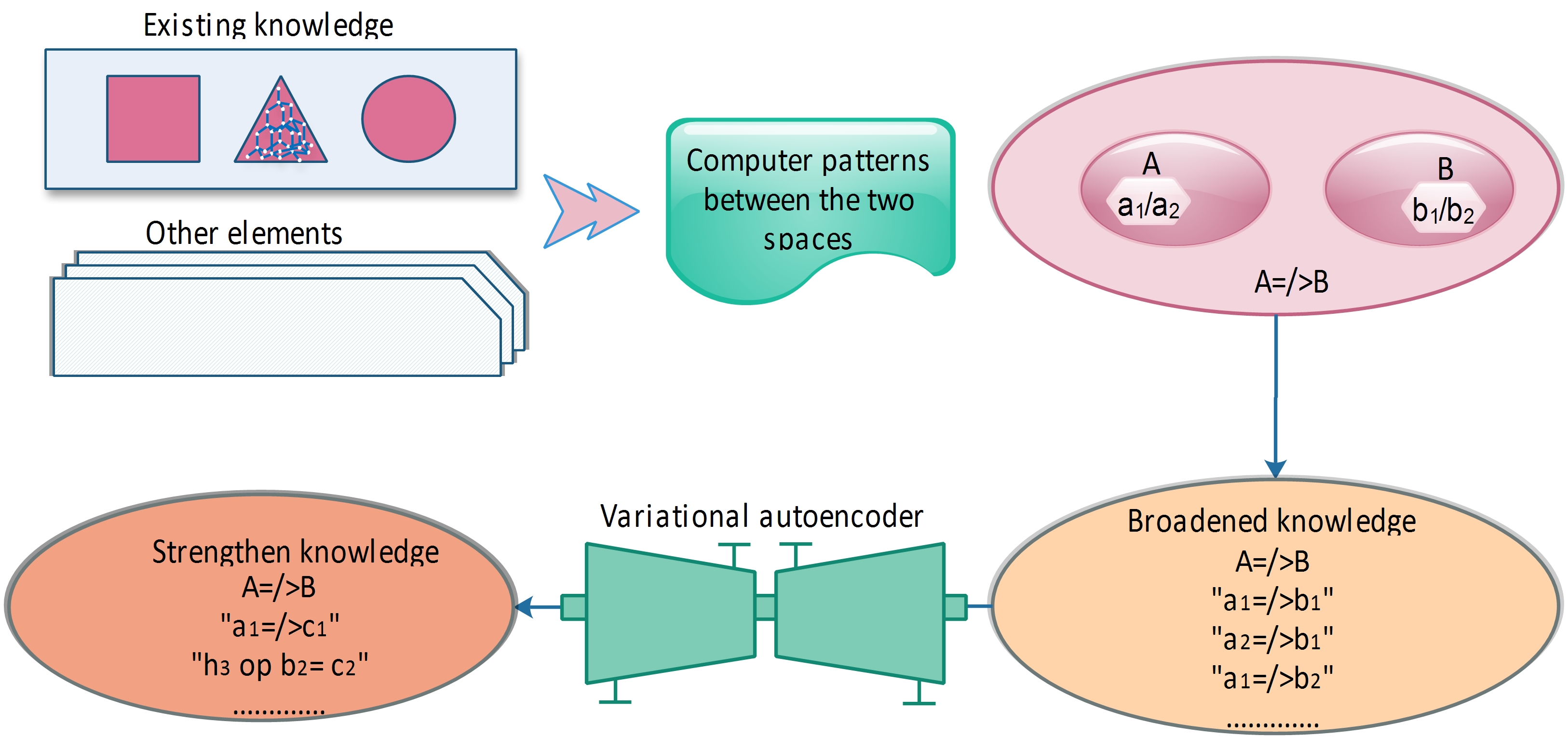}
		}
		\caption{Variational auto-encoding via NeSyL.}
		\label{fig:variational-based-NeSyL-integration}
	\end{figure}

	NeSyL can be incorporated into DL model such as VAE to reason via the enlightenment of the VAEs, the generation of the knowledge can be easily carried out. With the VAEs framework, we can build this knowledge-based data more easily in two ways. The first one is to use the logic inference, which can achieve a precise kind of knowledge, where the second one is to find the similarity and relation within knowledge, which might be approximated but also accurate to some extent. We can use the distribution to create knowledge. Using these two methods together can develop the initial knowledge (existing knowledge and a large amount of scattered but categorized elements), and reproduce a more abundant knowledge base. One of the generalized idea for the proposed model has been depicted in Figure~\ref{fig:variational-based-NeSyL-integration}.

	 In the first step, we use a small amount of existing knowledge about the visuals and elements (entities). The received existing knowledge and objects can be embedded and represented into corresponding spaces. A neural network extracts patterns from the entities like A and a1/a2, and their corresponding distributions. The key idea is to find a similar distribution of each class of entity and to enable referencing within a confidence coefficient (CC) such as 90\% CC level. For instance, the pattern $A > B$ can be used to expand our base knowledge into broadened knowledge: $a1 > b2$. The acquired broadened knowledge to facilitate logic reference processing and expansion with a larger amount of knowledge.
	 
	 In the next step, VAE model generates the broadened knowledge into strengthen broadened knowledge using variational encoding and decoding to map the similarities and the relations between the knowledge spaces. The model may create knowledge on its own by comparing the distributions of entities in the latent space. However, this is just a supposition without mathematical proof. We need some fundamental foundation with the mathematical tools and some experiments in the future study. After a round of iteration, the strengthened broadened knowledge can be generated either artificially or manually. Finally, the strengthened broadened knowledge can be used for the upcoming reasoning procedure within the symbolic AI.

	\subsubsection{NeSyL for Video Captioning and Internet of Things}
	In the study~\cite{akbari2020neuro}, neurosymbolic structure (via bias induction) has been proposed to caption visuals from videos by figuring out the key roles (predicates). The model receives videos coupled text descriptions and learns their cross-model relationship (via key attention-aware) to caption multi-model structures. To figure out the key roles in image and video captioning, further optimization is needed in roles division and selection in order to reason at the key role-level and sub-role level. This will allow humans to imitate captioning for a visual scenes corresponding to the query.  
	
	NeSyL based alerting and safety models~\cite{velik2010neurosymbolic} for different domains and applications such as homes, offices, stations, and buildings. Alerting model can be applied to critical environments such as safety, security, and forensic scenes, where symbol like gun, fire, knife, unusual movements and behaviors can be detected and captioned. Similarly, in the future, the autonomous system should be adapted to the alert model based on the human behaviors and emotions such as sleeping of a driver after long drive, monitoring driver health, and monitoring people involved in critical scenarios. 
	 
	\section{Discussion} AI faces a number of challenges, including emulating human learning, intuition, and reasoning abilities while also including interpretability into current AI \cite{hebb2005organization}. The basic plasticity principles that regulate the process by which environmental cues are converted into synaptic updates, however, are unclear. Many theories have been proposed, ranging from Hebbian-style processes that are physiologically reasonable but have yet to be demonstrated to tackle challenging real-world learning problems to a learning strategy that is successful but has multiple biologically dubious features. It is difficult to say whether precise measurements of activation patterns over time, synaptic qualities, or paired-neuron input-output correlations will allow for quantitatively accurate predictions of whether the findings are more compatible with one or another learning rule. This is a substantial difficulty by itself, because it is impossible to determine which patterns of neuronal changes originated from given learning rules on solely theoretical grounds without evaluating the overall network architecture and loss function of the learning system~\cite{nayebi2020identifying}. Rather than viewing NeSyL as universal improvements over deep learning, bias, robustness, and generalization, the architecture of NeSyL requires extensive research and international collaboration. NeSyL is one step closer to create a human brain mimic algorithm with two systems, which will open a door for novel discoveries and exciting applications.

	Neural networks have been rapidly expanding in recent years, with novel strategies and applications. However, challenges such as interpretability, explainability, robustness, safety, trust, and sensibility remain unsolved in neural network technologies, although they may unavoidably be addressed for critical applications. Attempts have been made to overcome the challenges in neural network computing, such as ML, by representing and embedding domain knowledge in terms of symbolic representations. Thus, neuro-symbolic learning (NeSyL) notion emerged, which incorporates aspects of symbolic representation and bringing common sense into neural networks (NeSyL). In domains where interpretability, reasoning, and explainability are crucial, such as video and image captioning, question-answering and reasoning, health informatics, and genomics, NeSyL has shown promising outcomes. This review presents a comprehensive survey on the state-of-the-art NeSyL approaches, their principles, advances in machine and deep learning algorithms, applications, and most importantly future perspectives. 
	
	\section{Conclusion} The widespread uses of neural networks are indispensable; nonetheless, the critical challenges such as black-box, interpretablility, and explainability limit its applications to certain scenarios including medical imaging and clinical practices. On the other hand, besides the white-box features and domain rich knowledge, symbolic learning has not yet advanced to the computational level of neural learning. Thus, it would be indispensable to infuse symbolics into neural learning as NeSyL to attain human intuitive features such as transperancy, interpretability, explainability, and reasoning in AI learning algorithms. In this study, we introduced relevant ideas, principles, and applications of NeSyL, so that a beginner reader can thoroughly understand the background knowledge before to conducting further research. We also revisited the existing CNN algorithms in novel directions infused with symbolic learning to pave the way for novel and robust NeSyL modalities. Most importantly, we introduced futuristic perspectives of NeSyL by advancing ML and DL modalities, and introducing viable NeSyL approaches for critical environments such as clinical practices, ocular diseases diagnosing, autonomous vehicles, health, and informatics.
	\appendices
	\section*{Acknowledgment}

	\ifCLASSOPTIONcaptionsoff
	\newpage
	\fi

	%\end{thebibliography}
	
	\bibliographystyle{ieeetr}

	%\bibliography{NeuroSymbL}

\end{document}